\pdfoutput=1
\documentclass[11pt]{article}

\usepackage[]{EACL2023}
\usepackage{times}
\usepackage{latexsym}
\usepackage{stmaryrd}
\usepackage{amsmath}
\usepackage{multirow}
\usepackage{bbm}
\usepackage{graphicx}
\usepackage{amssymb}
\usepackage{booktabs}
\usepackage{adjustbox}
\usepackage{xspace}
\usepackage{algpseudocode}
\usepackage{algorithm}
\usepackage{graphicx}
\usepackage{caption}
\usepackage{subcaption}
\usepackage{color, colortbl}
\usepackage{tablefootnote}
\usepackage{enumitem}  % More control
\usepackage{lipsum}
\usepackage{xspace}

\usepackage{hyperref}
\def\hyphenateAndTtWholeString #1{\xHyphenate#1$\wholeString\unskip}
\def\xHyphenate#1#2\wholeString {\if#1$%
    \else\transform{#1}%
    \takeTheRest#2\ofTheString\fi}
\def\takeTheRest#1\ofTheString\fi
{\fi \xHyphenate#1\wholeString}
\def\transform#1{\url{#1}\hskip 0pt plus 1pt}
\def\urlx #1{\href{#1}{\hyphenateAndTtWholeString{#1}}}

\usepackage[T1]{fontenc}
\usepackage[utf8]{inputenc}

\usepackage{microtype}

\newcommand{\datasetname}{NusaX}
\newcommand{\vquad}{\hspace{0.3em}} 
\newcommand{\ex}[1]{\textit{#1}\xspace} 
\newcommand{\equalsign}{\footnotemark[1]\hspace{0.1cm}}

\usepackage{inconsolata}

\setlist{topsep=1pt,itemsep=1pt,partopsep=1pt, parsep=1pt}

\title{\datasetname{}: Multilingual Parallel Sentiment Dataset \\ for 10 Indonesian Local Languages}

\author{
  Genta Indra Winata$^1$\thanks{\hspace{0.2cm}These authors contributed equally.}\hspace{0.12cm}, Alham Fikri Aji$^2$\equalsign, Samuel Cahyawijaya$^3$\equalsign, Rahmad Mahendra$^{4,5}$\equalsign, \\ \bf Fajri Koto$^{2,6}$\equalsign, \bf Ade Romadhony$^{5,7}$\equalsign,  Kemal Kurniawan$^{5,6}$\equalsign, David Moeljadi$^8$,\\ \bf Radityo Eko Prasojo$^9$, Pascale Fung$^3$, Timothy Baldwin$^{2,6}$, Jey Han Lau$^6$, \\ \bf Rico Sennrich$^{10}$, Sebastian Ruder$^{11}$\\
  $^1$Bloomberg \quad $^2$MBZUAI \quad $^3$HKUST \quad  $^4$Universitas Indonesia \quad $^{5}$INACL \\ $^6$The University of Melbourne \vquad $^7$Telkom University \vquad $^8$Kanda University of International Studies \\$^9$Kata.ai  \quad $^{10}$University of Zurich \quad $^{11}$Google Research \\
 }

\begin{document}
\maketitle
\begin{abstract}
Natural language processing (NLP) has significant impact on society via technologies such as machine translation and search engines. Despite its success, NLP technology is only widely available for high-resource languages such as English and Mandarin Chinese, and remains inaccessible to many languages due to the unavailability of data resources and benchmarks. In this work, we focus on developing resources for languages of Indonesia. Despite being the second most linguistically-diverse country, most languages in Indonesia are categorized as endangered and some are even extinct. We develop the first-ever parallel resource for 10 low-resource languages in Indonesia. Our resource includes sentiment and machine translation datasets, and bilingual lexicons. We provide extensive analysis, and describe challenges for creating such resources. Our hope is that this work will spark more NLP research on Indonesian and other underrepresented languages.
\end{abstract}

\section{Introduction}
% (Aji, Genta, Ade, Samcah, Rahmad)}

Indonesia is one of the most populous and linguistically-diverse countries in the world, with more than 700 languages spoken across the country~\cite{aji-etal-2022-one, ethnologue}. However, while many of these languages are spoken by millions of people they have received little attention from the NLP community. There are very few public datasets, preventing the global research community from exploring these languages. 
%On the other hand, research and progress from local institutions is scarce~\cite{purwarianti2007qa,Dinakaramani2014,ilmania2018aspect,ken2018entailment,saputri2018emotion,purwarianti2019improving,hoesen2018investigating,mahfuzh2019improving,kurniawan2018toward,ilmania2018aspect,azhar2019aspectcategorization,purwarianti2019improving,fern2019aspect,septi2019aspect}. Many of the locally developed resources are often not made publicly accessible. To spur progress in these underrepresented languages, it is crucial to develop medium to large scale datasets and benchmarks, such as IndoNLU~\cite{wilie2020indonlu}, IndoLEM~\cite{koto2020indolem}, IndoNLG~\cite{cahyawijaya-etal-2021-indonlg}, and IndoNLI~\cite{mahendra2021indonli}.
To this end, we introduce \textbf{\datasetname{}},\footnote{The dataset is released at~\url{https://github.com/IndoNLP/nusax}.} a high-quality multilingual parallel corpus that covers 10 local languages from Indonesia: Acehnese, Balinese, Banjarese, Buginese, Madurese, Minangkabau, Javanese, Ngaju, Sundanese, and Toba Batak.
% based on human translation and human-assisted quality assurance.

The \textbf{\datasetname{}} dataset was created by translating SmSA \cite{purwarianti2019improving} --- an existing Indonesian sentiment analysis dataset containing comments and reviews from the IndoNLU benchmark~\cite{wilie2020indonlu} --- using competent bilingual speakers, coupled with additional human-assisted quality assurance. Sentiment analysis is one of the most popular NLP tasks, 
%~\cite{arxiv-najjasenti} and has many practical uses in industry~\cite{cambria2017affective}. Sentiment analysis 
and has been explored in many applications in Indonesia, including presidential elections~\cite{ibrahim2015buzzer,budiharto2018prediction}, product reviews~\cite{fauzi2019word2vec}, stock forecasting~\cite{cakra2015stock,sagala2020stock}, and COVID-19 monitoring~\cite{nurdeni2021sentiment}. By translating an existing text, we additionally produce a parallel corpus, which is useful for building and evaluating translation systems. 
As we translate from a regional high-resource language (Indonesian), we ensure that the topics and entities reflected in the data are culturally relevant to the other languages, which is generally not the case when translating an English dataset \cite{conneau-etal-2018-xnli,ponti-etal-2020-xcopa}. 
% We present a parallel example of the NusaX corpus in Table~\ref{tab:nusaxexample}.
We apply the corpus to two downstream tasks: sentiment analysis and machine translation. 
% which is one of the most popular and developed NLP task in Indonesia. 
% This method enables us to synthetically generate more dataset on these languages, which can improve the progress of these languages further. 
We use the new benchmark to assess the performance of existing Indonesian language models (LMs), multilingual LMs, and classical machine learning methods. %in few-shot and full-data settings.

% Translating existing dataset has been used established as a standard data creation method~\cite{conneau2018xnli,lin2021xpersona}.

% We decide to opt sentiment analysis task due to . 
% Why sentiment:
% - Popular
% - Cost-effective
% - Use-case in Indonesia
% - 

% Why translating existing instead of labeling a new one?:
% - Retrieve data is hard
% - generate parallel corpus which is useful for MT task

% Austronesian -> Language Family
% Malayo-Polynesian, <Hijau>, <Biru>, <Biru Abu> -> Subgroup
% <Ungu> -> Language

% Contributions:
Our contributions are as follows:
\begin{itemize}
%JHL: anonymise link
    \item We propose \datasetname{}, the first high-quality human annotated parallel corpus in 10 languages from Indonesia, and corresponding parallel data in Indonesian and English, covering the tasks of sentiment analysis and machine translation.
    % \item We introduce datasets in these local languages covering two tasks: sentiment analysis and machine translation.
    % New Parallel Lexicon Resources.
    \item We provide an extensive evaluation of deep learning and classical NLP/machine learning methods on downstream tasks in few-shot and full-data settings.
    \item We conduct comprehensive analysis
    % regarding the similarity 
    of the languages under study both from linguistic and empirical perspectives, the cross-lingual transferability of existing monolingual and multilingual LMs, and an efficiency analysis of various methods for NLP tasks in extremely low-resource languages.
\end{itemize}

\section{Focus Languages}
% (David, Genta, Rahmad, Fajri)}

%FJ: if we can add a map of Indonesia, marking where these languages are spoken, it will be a great help. The reader is mostly non-Indonesian.

% ikutin MasakhaNER https://arxiv.org/ftp/arxiv/papers/2103/2103.11811.pdf

\begin{figure}[!t]
	\begin{subfigure}{1.0\linewidth}
		\centering
		\includegraphics[width=\linewidth]{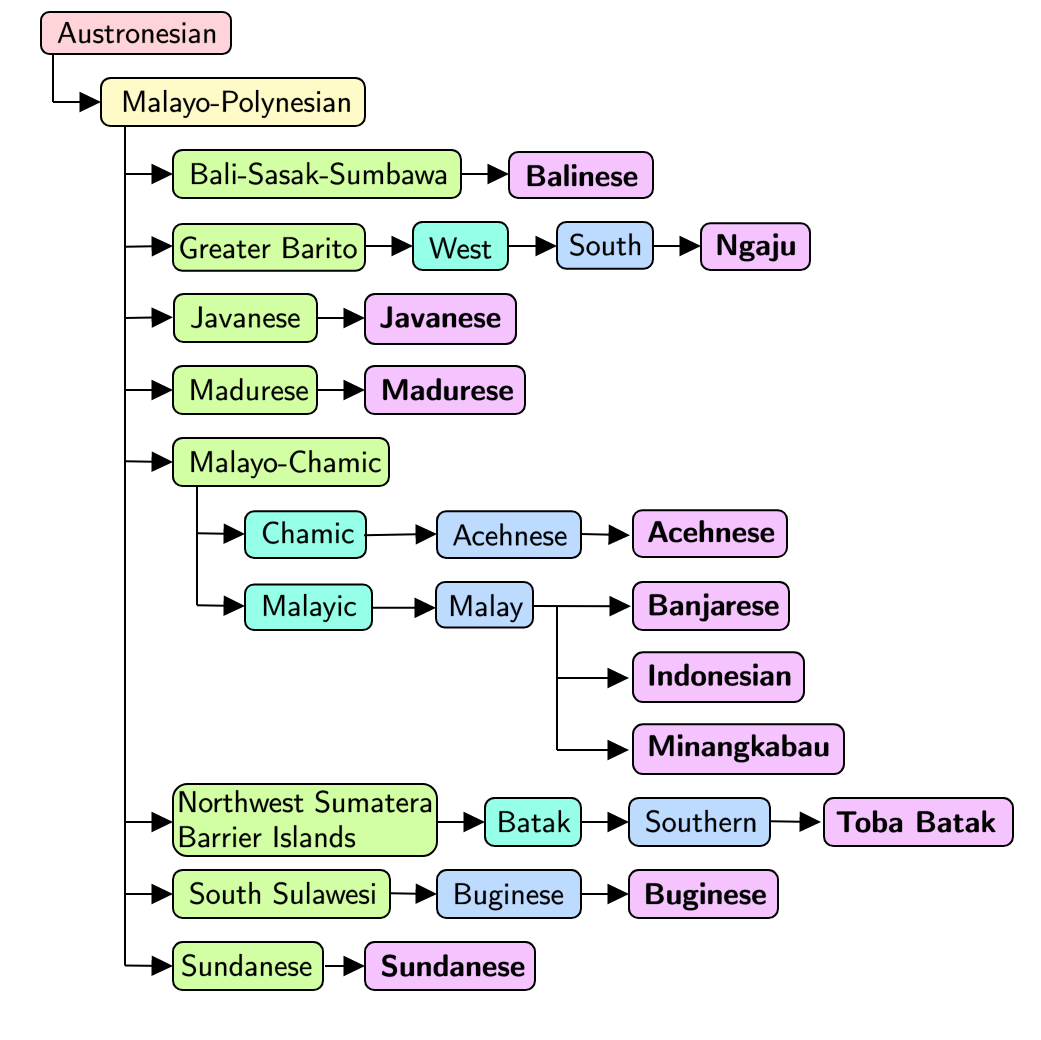} 
	\end{subfigure}
	\caption{Language taxonomy of the 10 focus languages and Indonesian, according to Ethnologue~\cite{ethnologue}. The color represents the language category level in the taxonomy. Purple denotes language, and other colors denote language family.}
	\label{fig:language-taxonomy}
\end{figure}

%We work on 10 local languages in Indonesia, namely \textbf{Acehnese} (4M speakers), \textbf{Balinese} (3M speakers), \textbf{Banjarese} (7M speakers), \textbf{Buginese} (5M speakers), \textbf{Madurese} (7M speakers), \textbf{Minangkabau} (6M speakers), \textbf{Javanese} (85M speakers), \textbf{Ngaju} (900k speakers), \textbf{Sundanese} (39M speakers), and \textbf{Toba Batak} (2M speakers). Most of these languages each have a population of over 2 million speakers~\cite{van2022writing} but are underrepresented in NLP research.
We work on 10 local languages in Indonesia: Acehnese, Balinese, Banjarese, Buginese, Madurese, Minangkabau, Javanese, Ngaju, Sundanese, and Toba Batak. Most of these languages have a population of over 2 million speakers~\cite{van2022writing,aji-etal-2022-one}, but are underrepresented in NLP research.
Figure~\ref{fig:language-taxonomy} shows the taxonomy of these languages and Indonesian. Geographically, these languages are spoken on different big islands in Indonesia, including Sumatra, Borneo, Java, Madura, and Sulawesi. 
%according to Ethnologue~\cite{ethnologue}. 
The languages belong to the Austronesian language family under the Malayo-Polynesian subgroup. %Table \ref{tab:language-data} provides additional information such as their status, number of L1 speakers, number of dialects, and writing systems.
While some of the covered languages are written in multiple scripts, we use the Latin script in NusaX, which has become predominant for all covered languages.

\textbf{Indonesian} (ind) is the national language of Indonesia based on the 1945 Constitution of the Republic of Indonesia (article 36). It is written in Latin script, and was developed from literary ``Classical Malay'' of the Riau-Johor sultanate \cite{sneddon2003}, with regional variants. Its lexical similarity to Standard Malay is over 80\%. 
%The word order is SVO. It has three optional noun classifiers. 
%It is non-tonal and has 19 consonants, 6 vowels, and 3 diphthongs. The stress is on the penultimate syllable. It has two social registers. 
It has a rich affixation system, including a variety of prefixes, suffixes, circumfixes, and reduplication. Most of the affixes in Indonesian are derivational~\cite{pisceldo-etal-2008-two}.

%is the national language of Indonesia in 1945 Constitution, Article 36 ~\cite{indonesia2002undang}. Its lexical similarity to Standard Malay is over 80\%. 
%The word order is SVO. %It has three optional noun classifiers. 
%It is non-tonal and has 19 consonants, 6 vowels, and 3 diphthongs. The stress is on the penultimate syllable. %It has two social registers. 
%It has a rich affixation system, including a variety of prefixes, suffixes, circumfixes, and reduplication. Most of the affixes in Indonesian are derivational.
%It is developed from literary `Classical Malay’ of the Riau-Johor sultanate \cite{sneddon2003} and has regional variants. It is written mainly in Latin script.

\textbf{Acehnese} (ace) is a language spoken mainly in the Aceh province. 
Although it is the de facto language of Aceh, language use is shifting to Indonesian in urban areas. 
Acehnese has features typical of the Mon-Khmer languages of mainland Southeast Asia, a result of its former status as part of the early Chamic dialect continuum on the coast of Vietnam. 
%It has at least ten contrasting vowels and as many distinct diphthongs, as well as voiceless aspirated stops and murmured voiced stops~\cite{blust2013austronesian}.
In addition to the large number of diphthongs, it has a high percentage of monosyllabic root morphemes. 
%Prefixes and infixes play an active role while suffixes are absent~\cite{Durie:1985}. It is of the `active’ or so-called `Split-S’ type: some intransitive verbs take arguments, which have the properties of `transitive subjects’ while others take arguments with the properties of ‘transitive objects’~\cite{durie1988preferred}.

\textbf{Balinese} (ban) is a language spoken mainly in the Bali province. %and in the West Nusa Tenggara province. 
It has three main dialects: Highland Balinese, Lowland Balinese, and Nusa Penida. Since the early 20th century, it has  mainly been written in the Latin script, but also has its own Balinese script.
The word order in Balinese is SVO. 
%It is non-tonal and has 17 consonant and 6 vowel phonemes. Stress is on the penultimate syllable. 
%Regarding patterns of verb affixation, Balinese is an `active' or `split-S' language: verbs with Undergoer-like subject arguments are marked in one way (with a `zero prefix'), while verbs with Actor-like subject arguments---intransitive or transitive---are marked in another (either with the nasal prefix `N-', or with `ma-'). 
Balinese has three sociolinguistic registers~\cite{arka2003balinese}.

\textbf{Banjarese} (bjn) is a language spoken in Kalimantan (Central, East, South, and West Kalimantan provinces). 
%It became a language of wider communication through trade in the market, in business, and in media. 
It is dominant in the South Kalimantan Province and is also growing rapidly in the Central and Eastern Kalimantan provinces. It has two main dialects: Kuala and Hulu. Although it is a Malayic language, it has many Javanese loanwords, probably acquired during the Majapahit period from the late thirteenth century until the fifteenth century~\cite{blust2013austronesian}. It has 73\% of lexical similarity with Indonesian and is written in Arabic and Latin scripts~\cite{ethnologue}.

\textbf{Buginese} (bug) is a language spoken mainly in the South Sulawesi, Southeast Sulawesi, Central Sulawesi, and West Sulawesi provinces. 
The word order is SVO. Verb affixes are used to mark persons. 
%It is non-tonal and has 19 consonant and 6 vowel phonemes. Stress is on the penultimate syllable. 
Historically, it was written in the Buginese script (derived from Brahmi script), but is mainly written in Latin script now~\cite{ethnologue}.
%In Buginese, the pronoun `I' has three forms: the independent form `iyya', the ergative form `-ka', and the absolutive form/clitic `u-'. 
Buginese employs sentence patterns, pronouns, and other terms to express politeness~\cite{weda2016syntactic}.

\textbf{Madurese} (mad) is a language spoken in the East Java province, mainly on Madura Island, south and west of Surabaya city, Bawean, Kangean, and Sapudi islands. 
% It is taught in primary and secondary schools.
It has vowel harmony, gemination, rich affixation, %three types of 
reduplication, and SVO basic word order \cite{davies2010grammar}.

\textbf{Minangkabau} (min) is a language spoken mainly in West Sumatra and other provinces on Sumatra Island such as Bengkulu and Riau. Although it is classified as Malay, it is not intelligible with Indonesian. 
%The word order is SVO written in Latin script. 
%It is taught in primary schools and written in Latin script.
Standard Minangkabau voice can be characterised as an Indonesian-type system, whereas colloquial Minangkabau voice is more effectively characterised as a Sundic-type system~\cite{crouch2009voice}.

\textbf{Javanese} (jav) is a language spoken mainly on Java Island.
%Banten, Central Java, East Java, and Yogyakarta provinces. 
It is the de facto language of provincial identity in central and eastern Java. The number of native Javanese speakers is greater than the number of Indonesian L1 speakers~\cite{ethnologue}. 
%At least 45\% of the total population of Indonesia are of Javanese descent or live in an area where Javanese is the dominant language.
%The word order is SVO. It has 21 consonants and 8 vowels. 
%Javanese differs from most other languages of western Indonesia in contrasting dental and retroflex stops, and in the feature of breathy voice or murmur as a phonetic property of its voiced obstruents. 
%Unlike most languages of the Philippines and western Indonesia, Javanese allows a number of word-initial consonant clusters. It has an elaborate system of speech levels~\cite{blust2013austronesian}.
Javanese consists of several regional dialects, which differ primarily in pronunciation and vocabulary. Javanese has an elaborate system of speech levels related to the relation of the speaker to the interlocutor that depend on social status, age, kinship distance, and familiarity~\cite{wedhawati2001tata}.
It used to be written in Javanese script, but since the 20th century has mostly been written in Latin script.

\textbf{Ngaju} (nij) is a language spoken in the Central Kalimantan province. It is widely used as a language 
% of wider communication 
for trade in much of Kalimantan, from the Barito to the Sampit River. 
%It is used in many domains (church, school, village-level government, market, etc.). 
It has various affixes and reduplication, and
%The active voice is marked by prefix `maN-' and the passive voice is marked by prefix `iN-'. 
its word order is similar to Indonesian. Pronouns have enclitic forms to mark possessors in a noun phrase or passive agents~\cite{UchiboriShibata1988}.

\textbf{Sundanese} (sun) is a language spoken mainly in the Banten and West Java provinces. It is the de facto language of provincial identity in western Java. The main dialects are Bogor (Krawang), Pringan, and Cirebon. 
%It is non-tonal and has 18 consonant and 7 vowel phonemes. The stress is on the penultimate syllable. 
It has elaborate coding of respect levels. It has been written in Latin script since the mid-19th century but was previously written in Arabic, Javanese, and Sundanese scripts.
Sundanese is a predominantly SVO language, and has voice marking and incorporates some (optional) actor-verb agreement, i.e., number and person~\cite{kurniawan2013sundanese}.

\textbf{Toba Batak} (bbc) is a language spoken in the North Sumatra province. %,  Samosir island, and east, south, and west of Lake Toba. 
Similarly to Acehnese, it is slowly being replaced by Indonesian in urban and migrant areas. It used to be written in the Batak script but is mainly written in Latin script now.
The Batak languages are verb-initial, and have verb systems reminiscent of Philippine languages, although they differ from them in many details~\cite{blust2013austronesian}.

\section{Data Construction}

%  Genta
Our data collection process consists of several steps. First, we take an existing dataset in a high-resource local language (Indonesian) as a base for expansion to the other ten languages, and ask human annotators to translate the text. To ensure the quality of the final translation, we run quality assurance with additional human annotators.
%JHL: need to make it clear that we're translating indonesian into 11 languages: 10 local languages and English

\subsection{Annotator Recruitment}
% (Genta, Rahmad)}
% Annotator recruitment. Brief information about annotators (detail in Appendix)
Eliciting or annotating data in underrepresented languages generally requires working with local language communities in order to identify competent bilingual speakers \cite{Nekoto2020}. In the Indonesian setting, this challenge is compounded by the fact that most languages have several dialects. %(cf. Table \ref{tab:language-data}).
%finding annotators for the collection of high-quality data in Indonesian languages is challenging due to the number of dialects of each language (cf. Table \ref{tab:language-data}).
As dialects in Indonesian languages may have significant differences in word usage and meaning~\cite{aji-etal-2022-one}, it is important to recruit annotators who speak the same or similar dialects to ensure that translations are mutually intelligible.

% Finding good annotators is a challenging process as we aim to collect high-quality data since two speakers of a language may understand different dialects. This poses a challenge when asking annotators to conduct quality assurance to make sure the quality assurance annotators understand the translations since two dialects in a language may have significant difference in word usage and meaning~\cite{aji-etal-2022-one}. 

In this work, we employ at least 2 expert annotators who are native speakers of each local language and Indonesian. %(see Appendix~\ref{app:statement}). 
To filter the recruited annotators, we first ask annotator candidates to translate three samples. We then conduct a peer review by asking whether they can understand the translations of other annotators for the same language, using the hired annotators as translators as well as translation validators. We also conducted 2 hours of training to introduce the user interface of the annotation system for selected workers. 
For English translations, we hire annotators based on their English proficiency test scores with an IELTS score $\ge$ 6.5 or TOEFL PBT score $\ge$ 600.
%If they both understand the translations, we decide that both annotators passed the qualification. Find the details in Appendix.

\subsection{Data Filtering and Sampling}
% (Genta, Samcah)}
% We select sentiment analysis because. 
We base our dataset on SmSA, the largest publicly available Indonesian sentiment analysis dataset from the \mbox{IndoNLU} benchmark~\cite{purwarianti2019improving,wilie2020indonlu}. SmSA is an expert-annotated sentence-level multi-domain sentiment analysis dataset consisting of more than 11,000 instances of comments and reviews collected from several online platforms such as Twitter, Zomato, and TripAdvisor. We filter the data to remove abusive language and personally-identifying information by manually inspecting all sentences. We randomly select 1,000 samples via stratified sampling for translation, ensuring that the label distribution is balanced.

\subsection{Human Translation}
% (Rahmad, Genta, Samcah)}

% We hire annotators to translate Indonesian text to local languages and to English. 
% More specifically, we find and select annotators who are fluent in speaking and writing Indonesian and a corresponding local language. 
%We verify the language proficiency of each annotator with two approaches: 1) a self-reporting questionnaire and 2) a test using a small-scale translation task consisting of 3 sentences in Indonesian. 
We instructed the annotators to retain the meaning of the text and to keep entities such as persons, organizations, locations, and time with no target language translation the same. Specifically, we instructed them to: (1) maintain the sentence's sentiment polarity; (2) preserve entities; and (3) maintain the complete information content of the original text.
% produce parallel sentences between the target language and Indonesian;

Initially, we asked the translators to maintain the typography. Most sentences from the original dataset are written in an informal tone, with non-standard spelling, e.g., elongated vowels and punctuation. When the sentence is translated into the target language, direct translation can sound unnatural. For example, translating the Indonesian word \ex{kangeeeen} (originally \ex{kangen}; en: \ex{miss}) to \ex{taragaaaak} (originally \ex{taragak}) in Minangkabau may sound unnatural. Similarly, the original sentence may also contain typos. Due to the difficulty of accurately assessing typographical consistency of translations, we removed this as a criterion. 
%We instruct annotators to use the Latin script as it is the most used writing system among locals.

\subsection{Human-Assisted Quality Assurance}
% (Genta, Aji, Samcah)}
% We choose quality assurance (QC) annotators that have similar dialects to the translation annotators. 
We conduct quality control (QC) between two annotators by having annotator A check the translations of annotator B, and vice versa. We include the corrected translations in our dataset. To ensure the quality assurance is performed well, we randomly perturb 5\% of the sentences by removing a random sequence of words. The quality assurance annotators are then expected to notice the perturbed sentences and fix them.

We analyze the quality assurance edits for Balinese, Sundanese, and Javanese, which are spoken by the authors of this paper. For each language, we randomly sample 100 translations that have been edited by a QC annotator. We classify edits as follows:

% \begin{itemize}
\noindent \textbf{Typos and Mechanics:} Edit that involves correcting typos, punctuation, casing, white spaces/dashes, and numerical formatting.

\noindent \textbf{Orthography:} Edit that changes the spelling of words due to orthographic variation in local languages without a standard orthography. The word sounds and means the same before and after editing, and both are used by natives. The QC annotator might feel that one writing variant is more natural/commonly used, and hence make this change.

\noindent \textbf{Translation:} The words used by the translator are still in Indonesian and the QC annotator translates them to the local language.

\noindent \textbf{Word edit:} The QC annotator paraphrases a word/phrase. This also includes adding/removing words and morpheme changes.

\noindent \textbf{Major changes:} Other edits that significantly alter the original translation.
% \end{itemize}

The results are shown in Table~\ref{tab:qc_edit}. Generally, word edits make up the majority of QC modifications, which involve replacing a word/phrase with a synonym or altering a morpheme slightly. In contrast, major changes are extremely rare. We also see changes to the orthography around 10\% of the time. Other types of edits vary between languages. Sundanese has significantly less typos compared to other languages, but a considerably higher number of translation edits. We suspect this is because code-switching with Indonesian happens regularly in Sundanese, which results in many Indonesian words being adopted despite the existence of equivalent Sundanese translations.

% We show the statistics of the quality control edits done by the annotators in Table~\ref{tab:qc_edit}. 
\begin{table}[!ht]
    \centering
    \resizebox{0.39\textwidth}{!}{
    \begin{tabular}{lccc}
    \toprule
        \textbf{Category} & \textbf{ban} & \textbf{sun} & \textbf{jav} \\
    \midrule
        Typos \& Mechanic & 31 & 14 & 42 \\
        Orthography & 14 & 6 & 12 \\
        Translation & 22 & 55 & 10 \\
        Word edit & 67 & 65 & 61 \\
        Major changes & 3 & 0 & 1\\
    \bottomrule
    \end{tabular}
    }
    \caption{Statistics of QC edits per category over 100 samples.}
    \label{tab:qc_edit}
\end{table}

\subsection{Bilingual Lexicon Creation}
% (Rahmad)}
Bilingual lexicons are useful for data augmentation \cite{wang-etal-2022-expanding} and evaluating cross-lingual representations \cite{artetxe-etal-2018-robust}. We select 400 words from an Indonesian lexicon\footnote{\url{https://github.com/andria009/IndonesianSentimentLexicon}} to be translated into the 10 local languages and English. For each language, we employ two annotators and ask them to translate the word into all possible lexemes. The translations from both annotators are combined. We obtain 800--1,600 word pairs for each of our 11 language pairs (from Indonesian to the remaining languages). We augment the bilingual lexicon with data from PanLex \cite{kamholz-etal-2014-panlex}.

\section{\datasetname{} Benchmark}
% (Aji, Rahmad, Samca, Genta)}

\subsection{Tasks}

We develop two tasks --- sentiment analysis and machine translation --- based on the datasets covering 12 languages, including Indonesian, English, and the 10 local languages. For the NusaX sentiment dataset, each language has the same label distribution and we show the label distribution of each dataset subset in Table~\ref{tab:label-distribution}. We maintain the label ratio in each dataset subset to ensure a similar distribution. More details of the dataset are provided in Appendix~\ref{app:statistics}.

% Sentiment analysis task is one of the most popular NLP tasks~\cite{arxiv-najjasenti}, especially in Indonesia~\cite{aji-etal-2022-one} and has many practical applications in the industry~\cite{cambria2017affective}. By translating existing text, we also produce a parallel corpus as a byproduct, which is useful for building translation systems. Translating existing dataset has also been used established as a standard data creation method~\cite{conneau2018xnli,lin2021xpersona}. 

% Why LID? untuk filtering pas scrapping Language technology for Scandinavian languages is in a nascent phase (e.g. Kirkedal et al. (2019)). One problem is acquiring enough text with which to train e.g. large language models. Good quality language ID is critical to this data sourcing, though leading models often confuse similar Nordic languages.

\subsubsection{Sentiment Analysis}
% (Samca, Genta)}

Sentiment analysis is an NLP task that aims to identify the sentiment of a given text document. The sentiment is commonly categorized into 3 classes: positive, negative, and neutral. We focus our dataset construction on sentiment analysis because it is one of the most widely explored tasks in Indonesia~\cite{aji-etal-2022-one} due to broad industrial relevance, such as for competitor and marketing analysis, and detection of unfavorable rumors
for risk management~\cite{socher2013recursive}. 
%JHL: missing ref
After translating 1,000 instances from the sentiment analysis dataset (SmSA), we  have a sentiment analysis dataset for each translated language. For each language, we split the dataset into 500 train, 100 validation, and 400 test examples. In total, our dataset contains 6,000 train, 1,200 validation, and 4,800 test instances across 12 languages (Indonesian, English and the 10 local languages). % We conduct experiments 5 times using different random seeds $\{0,...,4\}$ and report the average score of the evaluation metric.

% \subsubsection{Language Identification}
% % (Rahmad)}

% Language identification is a crucial tool for building large text corpora online. Large multilingual corpora, such as Common Crawl\footnote{\url{https://commoncrawl.org/}}, CC100~\cite{conneau-etal-2020-unsupervised}, and mC4~\cite{xue-etal-2021-mt5}, gather text from billions of web documents and categorise documents using a language identification tool. Existing language identification tools, however, focus on popular languages on the internet and tend to ignore underrepresented languages. 

% We introduce a dataset for language identification covering the 10 local Indonesian languages using our translated sentiment dataset. The size of the data is the same as our sentiment analysis dataset (6,000 train, 1,200 validation, and 4,800 test instances).

\subsubsection{Machine Translation}
% (Aji)}

Indonesia consists of 700+ languages covering three different language families~\cite{aji-etal-2022-one}. Despite its linguistic diversity, existing machine translation systems only cover a small fraction of Indonesian languages, mainly Indonesian (the national language), Sundanese, and Javanese. To broaden the coverage of existing machine translation systems for underrepresented local languages, we construct a machine translation dataset using our translated sentiment corpus, which results in a parallel corpus between all language pairs. In other words, we have 132 possible parallel corpora, each with 1,000 samples (500 train, 100 validation, and 400 test instances) which can be used to train machine translation models. Compared to many other MT evaluation datasets, our data is in the review domain and is not English-centric.

% \subsection{Evaluation Metrics}
% (Rahmad)}

%We utilize the macro-F1 score to measure the evaluation performance of sentiment analysis and language identification that are framed as classification tasks. For the machine translation task, we calculate the SacreBLEU score \cite{post-2018-call}.

%LID: Acc, Recall, Precision, Macro F1
%SA: Acc, Recall, Precision, Macro F1
%MT: BLEU, SemanticBLEU (BLEUScore/BLEURT)

\subsection{Baselines}
% (Aji, Rahmad, Genta)}

\subsubsection{Classical Machine Learning}

Classical machine learning approaches are still widely used by local Indonesian researchers and institutions due to their efficiency \cite{nityasya2021costs}. The trade-off between performance and compute cost is particularly important in situations with limited compute, which are common for low-resource languages.
We therefore use classical methods as baselines for our comparison. Namely, we use naive Bayes, SVM, and logistic regression for the classification tasks. For MT, we employ a naive baseline that copies the original Indonesian text, a dictionary-based substitution method using the bilingual lexicon, and a phrase-based MT system based on Moses~\cite{koehn2007moses}.
%JHL: citations for the MT baselines?
%JHL: need to also mention the copy baseline

% \paragraph{Classification Task}
% \paragraph{Generation Task}

\begin{table}[!t]
    \centering
     \resizebox{0.9\linewidth}{!}{
     \small
    \begin{tabular}{lccc}
    \toprule
        \textbf{Subset} & \textbf{Negative} & \textbf{Neutral} & \textbf{Positive} \\
    \midrule
        Train & 192 & 119 & 189 \\
        Valid & 38 & 24 & 38 \\
        Test & 153 & 96 & 151 \\
    \bottomrule
    \end{tabular}
    }
    \caption{Label distribution of NusaX Sentiment dataset.
    \label{tab:label-distribution}
    }
\end{table}

\begin{table*}[ht!]
\centering
\resizebox{0.98\textwidth}{!}{
\begin{tabular}{@{}llllllllllllll@{}}
\toprule
\textbf{Model} & ace & ban & bbc & bjn & bug & eng & ind & jav & mad & min & nij & sun & \textbf{avg} \\
 \midrule
%Classical & & & & & & & & & & & & \\
Naive Bayes & 72.5 & 72.6 & 73.0 & 71.9 & 73.7 & 76.5 & 73.1 & 69.4 & 66.8 & 73.2 & 68.8 & 71.9 & 72.0 \\
SVM & 75.7 & 75.3 & \textbf{76.7} & 74.8 & \textbf{77.2} & 75.0 & 78.7 & 71.3 & 73.8 & 76.7 & 75.1 & 74.3 & 75.4\\
LR & \textbf{77.4} & 76.3 & 76.3 & 75.0 & \textbf{77.2} & 75.9 & 74.7 & 73.7 & 74.7 & 74.8 & 73.4 & 75.8 & 75.4\\
\midrule
%Monolingual LM & & & & & & & & & & & & \\
IndoBERT$_{\textnormal{BASE}}$ & 75.4 & 74.8 & 70.0 & 83.1 & 73.9 & 79.5 & 90.0 & 81.7 & \textbf{77.8} & 82.5 & \textbf{75.8} & 77.5 & 78.5\\
IndoBERT$_{\textnormal{LARGE}}$ & 76.3 & \textbf{79.5} & 74.0 & 83.2 & 70.9 & 87.3 & 90.2 & \textbf{85.6} & 77.2 & 82.9 & \textbf{75.8} & 77.2 & \textbf{80.0} \\
IndoLEM$_{\textnormal{BASE}}$ & 72.6 & 65.4 & 61.7 & 71.2 & 66.9 & 71.2 & 87.6 & 74.5 & 71.8 & 68.9 & 69.3 & 71.7 & 71.1\\
% sundanese-roberta & 70.8 & 73.6 & 73.0 & 67.3 & 77.8 & & 75.1 & 72.2 & 68.9 & 73.4 & 80.7 & 64.7 \\
\midrule
%Multilingual LM & & & & & & & & & & & & \\
mBERT$_{\textnormal{BASE}}$ & 72.2 & 70.6 & 69.3 & 70.4 & 68.0 & 84.1 & 78.0 & 73.2 & 67.4 & 74.9 & 70.2 & 74.5 & 72.7\\
XLM-R$_{\textnormal{BASE}}$ & 73.9 & 72.8 & 62.3 & 76.6 & 66.6 & 90.8 & 88.4 & 78.9 & 69.7 & 79.1 & 75.0 & 80.1 & 76.2\\
XLM-R$_{\textnormal{LARGE}}$ & 75.9 & 77.1 & 65.5 & \textbf{86.3} & 70.0 & \textbf{92.6} & \textbf{91.6} & 84.2 & 74.9 & \textbf{83.1} & 73.3 & \textbf{86.0} & \textbf{80.0} \\

%\midrule
%XLM-R$_{\textnormal{LARGE}}$ (ALL) \\
\bottomrule
\end{tabular}
}
\caption{Sentiment analysis results in macro-F1 (\%). Models were trained and evaluated on each language.}
% \caption{Sentiment analysis results in macro-F1 (\%). Models were trained and evaluated on each language independently.}
\label{sentiment-results}
\end{table*}

%JHL: be consistent to use 'pretrained' or 'pre-trained'
\subsubsection{Pre-trained Local Language Models}

Recent developments in neural pre-trained LMs have brought substantial improvements in various NLP tasks. Despite the lack of resources in Indonesian and local languages, there have been some efforts in developing large pre-trained LMs for Indonesian and major local languages. IndoBERT~\cite{wilie2020indonlu} and SundaneseBERT~\cite{wongso2022pre} are two popular LMs for natural language understanding (NLU) tasks in Indonesian and Sundanese. IndoBART and IndoGPT have also been introduced for natural language generation (NLG) tasks in Indonesian, Sundanese, and Javanese \cite{cahyawijaya-etal-2021-indonlg}. We employ these LMs as baselines to assess their adaptability to other languages.
%JHL: IndoLEM isn't introduced here but used in table 3

\subsubsection{Massively Multilingual LMs}

We consider large pre-trained multilingual LMs to further understand their applicability to low-resource languages. Specifically, we experiment with mBERT~\cite{devlin2019bert} and XLM-R~\cite{conneau-etal-2020-unsupervised} for sentiment analysis, and mBART~\cite{liu-etal-2020-multilingual-denoising} and mT5~\cite{xue-etal-2021-mt5} for machine translation. We provide the hyper-parameters of all models in Appendix \ref{app:hyperparameters}.

% \section{Experiments}

% We conduct extensive experiments to set up baselines and provide in-depth analysis out of \datasetname{} dataset.

% \subsection{Sentiment Analysis (Samca, Aji)}

% For sentiment analysis, we 

% \subsubsection*{Cross-Lingual capability}

% \subsubsection*{Few-shot learning}

% \subsection{Language Identification (GENT, Samca)}

% \subsection{Machine Translation (Aji, GENTA)}

\section{Results}
\label{sec:results}

% The experiment details with all the hyperparameter settings are listed in Appendix X.

\subsection{Overall Results}

\paragraph{Sentiment Analysis}
% (Aji, Samca, Genta)}
    
\begin{table*}[!ht]
    \centering
    \resizebox{0.98\textwidth}{!}{
    \begin{tabular}{@{}lrrrrrrrrrrrr@{}}
    \toprule
    \multicolumn{13}{c}{\textbf{x $\rightarrow$ ind}} \\ \midrule
        \textbf{Model} & ace & ban & bbc & bjn & bug & eng & jav & mad & min & nij & sun & \textbf{avg}\\ 
    \midrule
        Copy & 5.88 & 9.99 & 4.28 & 15.99 & 3.44 & 0.57 & 9.29 & 5.11 & 18.10 & 7.51 & 9.24 & 8.13 \\
        Word Substitution & 7.33 & 12.30 & 5.02 & 16.17 & 3.52 & 1.67 & 17.34 & 7.89 & 24.17 & 12.07 & 15.38 & 11.17 \\ 
        PBSMT & \textbf{25.17} & \textbf{41.22} & \textbf{20.94} & 47.80 & \textbf{15.21} & 6.68 & 46.99 & \textbf{38.39} & \textbf{60.56} & 32.86 & 41.79 & 34.33 \\
        % From scratch & \\ 
        \midrule
        IndoGPT & 7.01 & 13.23 & 5.27 & 19.53 & 1.98 & 4.26 & 27.31 & 13.75 & 23.03 & 10.83 & 23.18 & 13.58\\
        % IndoBART & 18.27 & 32.64 & 13.91 & 38.74	& 9.38 & 3.70 & 37.20 & 28.39 & 49.37 & 27.32 & 30.97	\\
        IndoBARTv2 & 24.44 & 40.49 & 19.94 & \textbf{47.81} & 12.64 & 11.73 &	\textbf{50.64} & 36.10 & 58.38 & \textbf{33.50} & \textbf{45.96} & \textbf{34.69} \\
        mBART-50 & 18.45 & 34.23 & 17.43 & 41.73 & 10.87 & \textbf{17.92} & 39.66 & 32.11 & 59.66 & 29.84 & 35.19 & 30.64 \\
        mT5$_{\textnormal{BASE}}$ & 18.59 & 21.73 & 12.85 & 42.29 & 2.64 & 12.96 & 45.22 & 32.35 & 58.65 & 25.61 & 36.58	& 28.13 \\
    \bottomrule
    \end{tabular}
}
    \caption{Results of the machine translation task from other languages to Indonesian (x $\rightarrow$ ind) based on SacreBLEU.}
    \label{mt-results-lang-to-ind}
\end{table*}

% and for each task, we apply the same hyperparameter setting to ensure comparable results between all models. 

Table~\ref{sentiment-results} shows the sentiment analysis performance of various models across different local languages, trained and evaluated using data in the same language. Fine-tuned large LMs such as IndoBERT$_{\textnormal{LARGE}}$ and XLM-R$_{\textnormal{LARGE}}$ generally achieve the best performance. XLM-R models achieve strong performance on some languages, such as Indonesian (ind), Banjarese (bjn), English (eng), Javanese (jav), and Minangkabau (min). Many of these languages are included in XLM-R's pre-training data while others may benefit from positive transfer from related languages. 
% We conjecture that this might be because these languages are included among the supported languages in the multilingual model. 
% An exception is Banjarese, although it is very similar to Indonesian and Minangkabau. 
For instance, Banjarese is similar to Malay and Indonesian~\cite{nasution-2021-plan}, while Minangkabau shares some words and syntax with Indonesian~\cite{koto-koto-2020-towards}.
% positive transfer between closely related languages in the Malayo-Polynesian language family. 
IndoBERT models, despite only being pre-trained on Indonesian, also show good performance across some local languages, suggesting transferability from Indonesian to the local languages.

The classic approaches are surprisingly competitive with the neural methods, with logistic regression even outperforming IndoBERT$_{\textnormal{LARGE}}$ and XLM-R on Acehnese (ace), Buginese (bug), and Toba Batak (bbc). These results indicate that both Indonesian and multilingual pre-trained LMs cannot transfer well to these languages, which is supported by the fact that these languages are very distinct from Indonesian, Sundanese, Javanese, or Minangkabau --- the languages covered by IndoBERT and XLM-R.
%It also indicates that pre-trained models are not able to harness their knowledge from pre-training for these low-resource languages.

%  \begin{table}[!ht]
%     \centering
%     \resizebox{0.28\textwidth}{!}{
%     \begin{tabular}{lc} \toprule
%           Model & F1 \\ \midrule
% Naive Bayes & \textbf{99.44} \\
% SVM         & 98.42 \\
% LR          & 98.63 \\ \midrule 
% IndoBERT$_{\textnormal{BASE}}$ & 97.89 \\
% IndoBERT$_{\textnormal{LARGE}}$ & 97.87 \\
% IndoLEM$_{\textnormal{BASE}}$ & 97.10 \\ \midrule
% mBERT$_{\textnormal{BASE}}$ & 96.78 \\
% XLM-R$_{\textnormal{BASE}}$ & 96.83 \\
% XLM-R$_{\textnormal{LARGE}}$ & 97.09 \\ \bottomrule
% \end{tabular}
% }
%     \caption{Results of the language identification task.}
%     \label{lid-results}
% \end{table}

% hide

% \paragraph{Language Identification}
% (Aji, Samca, Genta)}

% Table~\ref{lid-results} shows the language identification performance of various models. Both pre-trained language models and classical machine learning approaches perform well on the language identification task. The classical models marginally outperform the deep learning models; Naive Bayes in particular, has the best performance. This suggests that language identification is an arguably simple problem that can be modelled with linear models. % logistic regression. 
%However, note that the models were trained and tested in-domain. It is thus necessary to further test them on a more general domain to assess their true generalization ability.

\paragraph{Machine Translation} 
% (Aji, Genta)}

%JHL: need to use consistent typefont for method (here we see copy is in texttt, but not in the table or other places)
We show the results on machine translation in
Table~\ref{mt-results-lang-to-ind} (x $\rightarrow$ ind) based on SacreBLEU~\cite{post-2018-call}. As some local languages are similar to Indonesian, we observe that the \texttt{Copy} baseline (which does not do any translation) performs quite well. Minangkabau (min) and Banjarese (bjn) achieve high BLEU without any translation despite not being included in the LM pre-training data, due to their similarity with Indonesian~\cite{koto-koto-2020-towards, nasution-2021-plan}.
Since these local languages share grammatical structure with Indonesian, dictionary-based word substitution yields a reasonable improvement.

\begin{table}[t!]
    \centering
    \resizebox{0.49\textwidth}{!}{
    \begin{tabular}{lcccc}
        \toprule
         \textbf{Model} & \multicolumn{4}{c}{\textbf{avg. SacreBLEU}} \\
         & ind $\shortrightarrow$ x & x $\shortrightarrow$ ind & eng $\shortrightarrow$ x & x $\shortrightarrow$ eng \\
         \midrule
          PBSMT &  \textbf{28.72} & 34.33 & 4.56 & 5.84 \\
          IndoBARTv2  & 28.21 & \textbf{34.69} & 6.36 & \textbf{7.46} \\
          mBART-50 & 24.69 & 30.64 & \textbf{7.20} & 6.45 \\
         \bottomrule
    \end{tabular}
    }
    \caption{MT performance from / to Indonesian compared to from / to English.}
    \label{tab:en-x}
\end{table}

Both PBSMT and fine-tuned LMs reach encouraging performance levels despite the limited training data, which we again attribute to the target languages' similarity to Indonesian. In contrast, the performance for translating Indonesian languages from/to English is extremely poor as shown in Table~\ref{tab:en-x}, demonstrating the importance of non-English-centric translation. Overall, we observe good translation performance across local languages. Thus, there is an opportunity to utilize translation models to create new synthetic datasets in local languages via translation from a related high-resource language, not only for Indonesian local languages but also other underrepresented languages. 
However, note that even for language pairs where the SacreBLEU score is very high, we observe translation deficiencies stemming from the small amount of training data: rare words may just be copied with PBSMT, and mistranslated with NMT. 

Similar effects are also observed for (ind $\rightarrow$ x) translation, as shown in Table~\ref{mt-results-ind-to-lang}. Similar to (x $\rightarrow$ ind) translations, we observe that the \texttt{Copy} baseline performs quite well on Minangkabau (min) and Banjarese (bjn) due to their similarity with Indonesian~\cite{koto-koto-2020-towards,nasution-2021-plan}. Dictionary-based word substitution also yields a reasonable improvement especially for Javanese (jav), Minangkabau (min), and Sundanese (sun) due to the high similarity of the grammatical structure with Indonesian. PBSMT and fine-tuned IndoBARTv2 models achieve the best scores over multiple local languages despite the limited training data, which is also attributed to the target languages' similarity to Indonesian. 

\begin{table*}[!t]
    \centering
    \resizebox{0.99\textwidth}{!}{
    \begin{tabular}{@{}lrrrrrrrrrrrr@{}}
    \toprule
    \multicolumn{13}{c}{\textbf{ind $\rightarrow$ x}} \\ \midrule
        Model & ace & ban & bbc & bjn & bug & eng & jav & mad & min & nij & sun & \textbf{avg} \\ 
    \midrule
        Copy & 5.89 & 10.00 & 4.28 & 15.99 & 3.45 & 0.56 & 9.29 & 5.11 & 18.10 & 7.52 & 9.24 & 8.13 \\ 
        Word Substitution & 7.60 & 10.31 & 5.99 & 17.51 & 3.57 & 0.76 & 14.75 & 7.58 & 22.34 & 9.76 & 12.38 & 10.23 \\ 
        PBSMT & \textbf{20.47} & 26.48 & 18.18 & \textbf{42.08} & 10.84 & 7.73 & 39.08 & \textbf{33.26} & \textbf{52.21} & \textbf{29.58} & 36.04 & \textbf{28.72} \\
        % From scratch & \\ 
        \midrule
        IndoGPT & 9.60 & 14.17 & 8.20 & 22.23 & 5.18 & 5.89 & 24.05 & 14.44 & 26.95 & 17.56 & 23.15 & 15.58 \\
        % IndoBART & 15.08 & 22.42 & 13.09 & 35.73 & 6.60 & 4.42 & 32.92 & 23.47 & 40.21 & 23.34 & 29.74 \\
        IndoBARTv2 & 19.21 & \textbf{27.08} & \textbf{18.41} & 40.03 & \textbf{11.06} & \textbf{11.53} & \textbf{39.97} & 28.95 & 48.48 & 27.11 & \textbf{38.46} & 28.21 \\
        mBART-50 & 17.21 & 22.67 & 17.79 & 34.26 & 10.78 & 3.90 & 35.33 & 28.63 & 43.87 & 25.91 & 31.21 & 24.69\\
        mT5$_{\textnormal{BASE}}$ & 14.79 & 18.07 & 18.22 & 38.64 & 6.68 & 11.21 & 33.48 & 0.96 & 45.84 & 13.59 & 33.79 & 21.39 \\
    \bottomrule
    \end{tabular}
}
    \caption{Results of the machine translation task from Indonesian to other languages (ind $\rightarrow$ x) in SacreBLEU.}
    \label{mt-results-ind-to-lang}
\end{table*}

% This provides additional motivation for the creation of parallel data for local languages.

\subsection{Cross-lingual Capability of LMs}
\label{sec:multilingual}
% (Samca, Genta, Aji, Rahmad)}

From a linguistic perspective, local languages in Indonesia share similarities according to language family. Many local languages share a similar grammatical structure and have some vocabulary overlap. Following prior work that demonstrates positive transfer between closely-related languages~\cite{cahyawijaya-etal-2021-indonlg,hu2020xtreme,aji2020neural,khanuja2020gluecos,winata2021language,winata2022cross}, we analyze the transferability between closely-related languages in the Malayo-Polynesian language family.

\begin{figure}[!t]
    \centering
    \includegraphics[width=\linewidth]{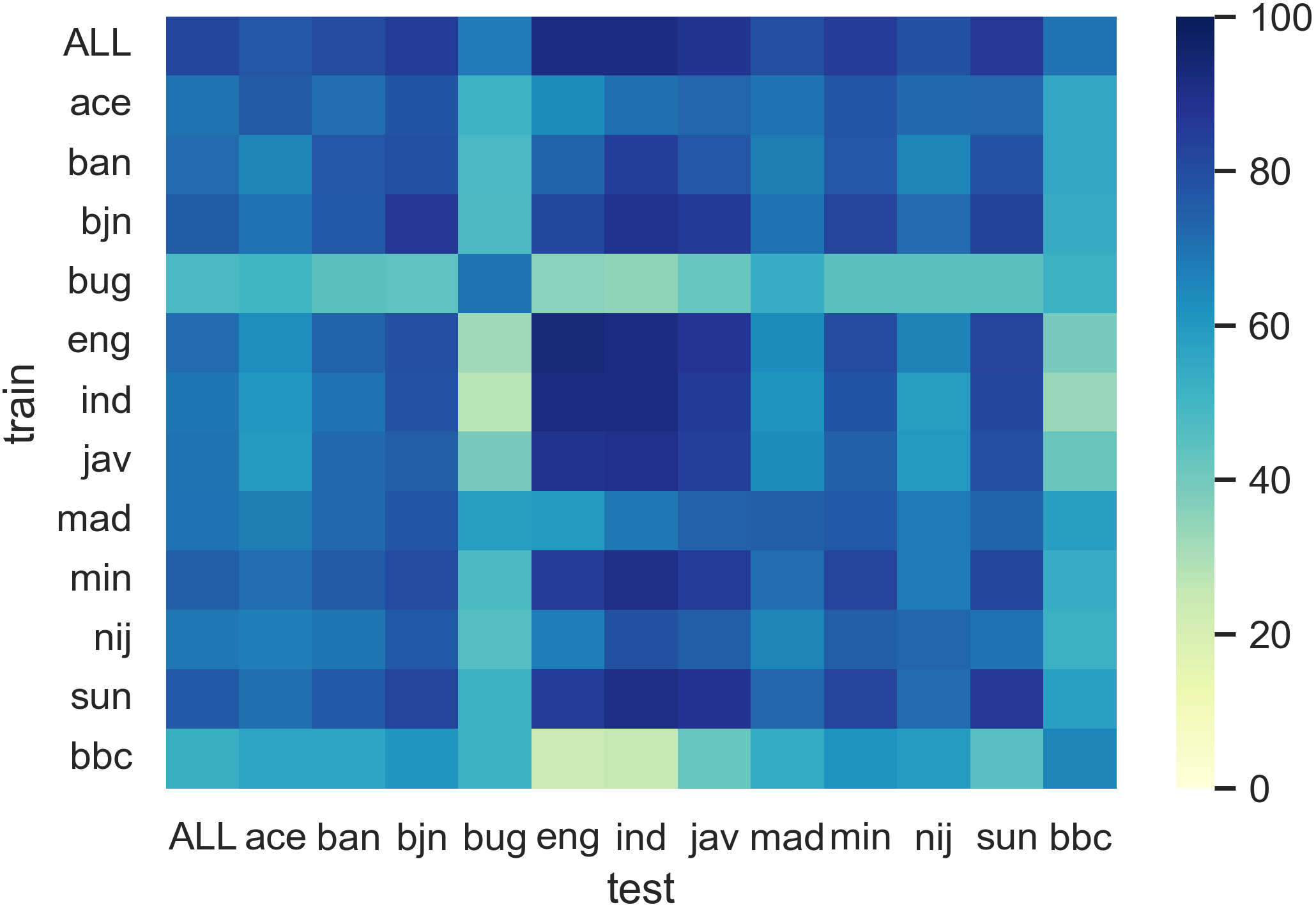}
    \caption{Zero-shot cross-lingual results for the sentiment analysis task with XLM-R$_{\textnormal{LARGE}}$. The model is trained on the language indicated on the $x$-axis and evaluated on all languages.}
    \label{fig:cross-lingual-results-sentiment}
\end{figure}

%JHL: need to explain what is the zero-shot cross-lingual setting (from the figure caption it says it's trained on one language and evaluated on other languages)
Empirically, we show the cross-lingual capability of the best performing model (XLM-R$_{\text{LARGE}}$) in the zero-shot cross-lingual setting for sentiment analysis. The heatmap is shown in Figure~\ref{fig:cross-lingual-results-sentiment}. In general, most languages, except for Buginese (bug) and Toba Batak (bbc), can be used effectively as the source language, reaching $\sim$70--75\% F1 on average, compared to an average of 80\% F1 in the monolingual setting (cf.\ XLM-R$_\text{LARGE}$ in Table~\ref{sentiment-results}). This empirical result aligns with the fact that both Buginese (bug) and Toba Batak (bbc) have very low vocabulary overlap with Indonesian (cf.\ Copy in Tables \ref{mt-results-lang-to-ind} and \ref{mt-results-ind-to-lang}). Interestingly, despite coming from a completely different language family, English can also be effectively used as the source language for all 10 local languages, likely due to its prevalence during pre-training.
% We present the generalizability of multilingual models in detail in Section~\ref{sec:multilingual}.
%JHL: missing ref

% As expected, multilingual training using all languages provides the best result, reaching an average of 81.6\% F1 across all languages. This is better than training the model on each individual language, which yields around 79.0\% F1 

% across all languages 

% Despite having lower score than the multilingual training, we observe a high F1 score ($\sim$79\%) in zero-shot cross lingual sentiment analysis task for the model trained on individual language.

%JHL: are we not just repeating the 

% In general, most languages can be used effectively as the source language, aside from Buginese (bug) and Toba Batak (bbc). The two languages possess a very low vocabulary overlap with Indonesian~\cite{aji-etal-2022-one}. 
These results demonstrate that we can take advantage of language similarity by transferring knowledge from Indonesian and other local languages to perform zero-shot or few-shot classification in closely-related languages. New datasets for underrepresented languages that are closely related to high-resource languages thus do not necessarily need to be large, which make the development of NLP datasets in low-resource languages more affordable than may initially  appear to be the case.

\subsection{Multilingual Capability}

\begin{table}[!t]
    \centering
    \resizebox{0.9\linewidth}{!}{
    \small
    \begin{tabular}{lccc}
    \toprule
	\textbf{Language} &	\textbf{Single}	&	\textbf{Multi}	&	\textbf{LOLO}	\\
	\midrule
Acehnese	&	75.9	&	76.96	&	75.79	\\
Balinese	&	77.1	&	80.13	&	77.83	\\
Banjarese	&	86.3	&	84.85	&	82.68	\\
Buginese	&	70.0	&	67.86	&	63.67	\\
English	&	92.6	&	91.05	&	89.88	\\
Indonesian	&	91.6	&	91.13	&	90.62	\\
Javanese	&	84.2	&	88.19	&	87.39	\\
Madurese	&	74.9	&	79.41	&	78.52	\\
Minangkabau	&	83.1	&	85.29	&	84.45	\\
Ngaju	&	73.3	&	78.82	&	76.31	\\
Sundanese	&	86.0	&	86.02	&	84.41	\\
Toba batak	&	65.5	&	70.00	&	68.76	\\

\midrule
Average	&	80.04	&	81.64	&	80.03	\\
\bottomrule
\end{tabular}
}
\caption{Sentiment analysis results for macro-F1 (\%) of XLM-R$_{\textnormal{LARGE}}$ in the multilingual setting.}
\label{tab:muti-sent}
\end{table}

We explore training multilingual models, as most Indonesian local languages share similarities. For sentiment analysis, we concatenate the training data of all languages. Additionally, we also explore Leave-One-Language-Out (LOLO), where we train on all data except for the test language. The LOLO setting arguably reflects the most realistic scenario where we do not have training data for a particular language, but we do have access to data in other local languages. The multilingual results for sentiment analysis are shown in Table~\ref{tab:muti-sent}. Multilingual training outperforms monolingual training, while LOLO matches the performance of training on target language data. Related language data is thus often sufficient for good cross-lingual results.

%in most languages, aside for Buginese (bug) and Toba Batak (bbc), which we argue because of ....?

%- Cross-lingual test results
%- Model that performs best and why?
%- Multilingual vs Monolingual models capability

% \subsection{Language Similarity of Local Languages}

% not for arxiv, submission only
% \subsection{Multilingual Learning}

% \subsection{Performance-Efficiency in Extremely Low-Resource Languages (Aji, CAH)}

%\section{Discussion}

\section{Data Collection Challenges}

In this section, we discuss challenges faced during data collection.

\paragraph{Finding annotators} We found collecting the NusaX dataset challenging. First of all, finding local language-speaking annotators is not easy, and popular platforms such as MTurk do not support these languages. 
% (it's not as easy as looking in platforms like MTurk).
Instead, we looked for annotators through local Indonesian networks and forums, such as the INACL forum, local campus forums, or the Indonesian polyglot community. We intended to cover as many local languages as possible, but based on the available annotators, only the 10 languages presented in this paper were possible, as we needed at least 2 annotators for each language. Searching for annotators online is not easy, due to disparities in Internet penetration in different parts of Indonesia. Hence, we might not reach potential annotators through online communities alone. However, holding an in-person workshop for data collection is also not practical; Indonesia is an archipelago and traveling between islands is costly. 
Similar challenges occur in many other regions, including Africa and South America.

%  Local languages in Indonesia are rich in dialects, and the lexical variation between dialects can be very different, despite being the same language. In addition, the annotators for each language must speak the same dialect.

\paragraph{Communication with annotators} 
Communication between the authors and annotators was done through WhatsApp, as the most popular communication tool in Indonesian~\cite{mulyono2021application}. Annotation was conducted through spreadsheets. We found that some of the annotators use mobile apps instead of a desktop for annotation. Their reasons include ease of use, % due to not having 
no access to a laptop, and better keyboard support for typing diacritics. In the most extreme case, one annotator printed out the sheet and performed the annotation on paper, then took a picture of the paper and sent it back to us. We found some annotators to be difficult to contact, due to other commitments such as college or work. Some of them were not responsive and had to be replaced by new annotators.

% \subsection{Data Size}

% Given the good performance in MT and sentiment analysis with only a limited dataset size, one of the directions for further research is to expand more languages. In addition, exploration for other tasks and domains will also be very interesting. We also look at the cross-lingual zero-shot capability in sentiment analysis. Further exploration would also be interesting, for example in the direction of zero-shot cross-task for local languages. Given the challenges in data collection, exploration to collect data sources in local languages in Indonesia may also be a direction for future research.

% \subsection{Broader Impact}

% NusaX enables NLP research for under-represented languages. NusaX can be used as a testbed for transfer learning or few-shot learning methods that take advantage of similarities between languages. The resources provided also bring opportunities for local researches or communities to conduct research on local languages domain, since currently finding publicly available local language datasets is one of the main challenges they face.

% NusaX opens up the possibility for future research to focus on covering more local languages, and additionally, further advancement to other tasks and domains. Our study on cross-lingual transfer enables further exploration on cross-lingual zero-shot learning for more diverse tasks in local languages. Our guideline and discussion on data collection may also enlighten future research work to further refine a more efficient high quality data collection guideline for extremely low-resource languages.

\section{Related Work} 
% (Ade, Rahmad, Genta)}

% Model: IndoNLU, IndoNLG,IndoLEM, IndoBERTTweet, SundaneseBert,  mBERT, XLM, XLM-R
% Benchmark: IndoNLU, IndoNLG, IndoLEM, GLUE, CLUE, KLUE
% Low-Resource: 1C700+L, Something else.
% Dataset initiative (local): IndoNLI, ...

\paragraph{Multilingual Parallel Corpora}
% (Ade, Rahmad, Genta)}
Several multilingual parallel corpora have been developed to support studies on machine translation such as GCP~\cite{imamura2018multilingual}, Leipzig~\cite{goldhahn-etal-2012-building}, JRC Acquis \cite{steinberger2006jrc}, TUFS Asian Language Parallel~\cite{nomoto2018tufs}, Intercorp \cite{ek2012case}, DARPA LORELEI \cite{strassel2016lorelei}, Asian Language Treebank \cite{riza2016introduction}, FLORES \cite{guzman-etal-2019-flores}, the Bible Parallel Corpus \cite{resnik1999bible, black2019cmu}, JW-300~\cite{agic-vulic-2019-jw300}, BiToD~\cite{lin2021bitod}, and WikiMatrix~\cite{schwenk-etal-2021-wikimatrix}. 
%The languages covered in the GCP corpus are Japanese, English, Spanish, French, and several Asian languages including Indonesian. While the GCP corpus covers sentence pairs in tourism related domains, JRC Acquis contains mostly legal documents in 20 official EU languages. Similar to the GCP corpus, TUFS Asian Corpus (TALPCo) covers Japanese sentences and their translations to English, Burmese, Malay, and Indonesian. Intercorp is a multilingual corpus based on Czech covering 27 other languages. 
% DARPA LORELEI and the Asian Language Treebank cover several low resources languages, while FLORES contains sentences in two low resource languages, parallel with English. 
% Building high-quality parallel corpora is expensive and time-consuming, since it requires translation by native or fluent translators.
\citet{guzman-etal-2019-flores} describe the procedure to generate high-quality translations as part of FLORES.
% The translation quality checks consist of two types of filtering: automatic and manual. 
Similar to FLORES, we also conducted QC of the translations.

\paragraph{Emerging Language Benchmarks} 
% (Rahmad, Genta)}
Recently, % datasets and 
benchmarks
in underrepresented languages have emerged, such as MasakhaNER \cite{tacl_masakhaner}, AmericasNLI \cite{ebrahimi-etal-2022-americasnli}, PMIndia \cite{haddow2020pmindia}, Samanantar \cite{tacl_samanantar}, and NaijaSenti \cite{muhammad-etal-2022-naijasenti}. 
% MasakhaNER covers ten African languages, while AmericasNLI covers ten indigenous languages of Americas and NaijaSenti covers four Nigerian languages. 
Particularly, for Indonesian languages, NLP benchmarks have been developed such as  IndoNLU~\cite{wilie2020indonlu}, IndoLEM~\cite{koto2020indolem}, IndoNLG~\cite{cahyawijaya-etal-2021-indonlg}, IndoNLI~\cite{mahendra-etal-2021-indonli}, and English--Indonesian machine translation~\cite{guntara2020benchmarking}.

%These benchmarks mostly focus on Indonesian, except for IndoNLG, which also includes Javanese and Sundanese.
% In recent years, the development of multilingual parallel corpus also involves local languages, such as PMIndia \cite{haddow2020pmindia} and Samanantar \cite{tacl_samanantar}. 
%PMIndia covers 13 Indic languages where each language pair (English with one of languages of India) contains a maximum of 56,000 sentence pairs, sourced from the website of India's prime minister. Samanantar covers 11 Indic languages, with a total of 49.7 million sentence pairs where one source corpus is PMIndia. 

%Researches on multilingual language model and also multilingual benchmark dataset creation has been emerged lately, including initiatives on multilingual low resource dataset, such as MasakhaNER \cite{tacl_masakhaner}, AmericasNLI \cite{americasnli}, and NaijaSenti \cite{arxiv-najjasenti}. MasakhaNER covers ten African languages, while AmericasNLI covers ten indigenous languages of Americas and NaijaSenti covers four Nigerian languages. Our dataset is a sentiment dataset, similar to NaijaSenti, however NaijaSenti is not a parallel dataset, the main work is on annotating the sentiment to the collected document.  Other task-specific benchmarks include En-Id machine translation~\cite{guntara2020benchmarking}, paraphrasing~\cite{aji2021paracotta}, and low-cost sequence classification and labeling in~\cite{nityasya2021costs,nityasya2022student}. 

\paragraph{Datasets for Indonesian Local Languages}
% (Rahmad)}
%Numerous multilingual benchmarks have established to cater multilingual NLP research. 
Only a limited number of labeled datasets exist for local languages in Indonesia. WikiAnn~\cite{pan-etal-2017-cross} --- a weakly-supervised named entity recognition dataset --- covers Acehnese, Javanese, Minangkabau, and Sundanese. \citet{Putri_etal_2021_abusive} built a multilingual dataset for abusive language and hate speech detection involving Javanese, Sundanese, Madurese, Minangkabau, and Musi languages. \citet{sakti2013towards} constructed speech corpora for Javanese, Sundanese, Balinese, and Toba Batak. Few datasets exist for individual languages, e.g., sentiment analysis and machine translation in Minangkabau~\cite{koto-koto-2020-towards} and emotion classification in Sundanese~\cite{putra2020sundanese}. Finally, some datasets focus on colloquial Indonesian mixed with local languages in the scope of morphological analysis~\cite{wibowo2021indocollex} and style transfer~\cite{wibowo2020semi}.

\section{Conclusion}
% (Genta)}
In this paper, we propose \datasetname{}, the first parallel corpus for 10 low-resource Indonesian languages. We create a new benchmark for sentiment analysis and machine translation in zero-shot and full-data settings. We present a comprehensive analysis of the language similarity of these languages from both linguistic and empirical perspectives by assessing the cross-lingual transferability of existing Indonesian and multilingual pre-trained models.

We hope NusaX can enable NLP research for under-represented languages, and can be used as a testbed for adaptation or few-shot learning methods that take advantage of similarities between languages. 
%NusaX also bring opportunities for local researches or communities to conduct research on domains relevant to local languages, since currently finding publicly available local language datasets is one of the main challenges they face.
NusaX opens up the possibility for future research that focuses on covering more local languages, and additionally, further extension to other tasks and domains. Our study on cross-lingual transfer enables further exploration on cross-lingual zero-shot learning for more diverse tasks in local languages. Our guidelines and discussion of data collection issues may also motivate future work on more efficient high-quality data collection for extremely low-resource languages.

% \section*{Broader Impact}

% NusaX enables NLP research for under-represented languages. NusaX can be used as a testbed for transfer learning or few-shot learning methods that take advantage of similarities between languages. The resources provided also bring opportunities for local researches or communities to conduct research on local languages domain, since currently finding publicly available local language datasets is one of the main challenges they face.

% NusaX opens up the possibility for future research to focus on covering more local languages, and additionally, further advancement to other tasks and domains. Our study on cross-lingual transfer enables further exploration on cross-lingual zero-shot learning for more diverse tasks in local languages. Our guideline and discussion on data collection may also enlighten future research work to further refine a more efficient high quality data collection guideline for extremely low-resource languages.

% \newpage

\section*{Acknowledgments}
We thank Dea Adhista and all annotators who helped us in building the corpus. We are grateful to Alexander Gutkin and Xinyu Hua for feedback on a draft of this manuscript. This work has been partially funded
by Kata.ai (001/SD/YGI-NLP/1/2022) and PF20-43679 Hong Kong PhD Fellowship Scheme, Research Grant Council, Hong Kong.

\section*{Limitations}

We created data for low-resource languages, which increases the accessibility of NLP research for marginalized communities. However, we were only able to cover 10 languages with only 1000 samples each, due to cost and the number of available annotators. This dataset has limited domain coverage and may also contain biases towards certain groups or entities. We tried our best to eliminate negative biases based on a manual inspection of the data. As our dataset was translated, there may be some translationese artifacts in the resulting corpus.
% We argue
% that further study on the potential bias and the translationese effect of our dataset is needed.
We invited annotators based on their fluency level on a particular language. However, the fluency level is self-declared, and there is no mechanism to verify it, except for several languages that are spoken by authors of this paper. The dialect used in the dataset also depends on the annotator, for languages with multiple dialects. 

% Our best-performing systems are based on large LMs, which can be costly to train and deploy. Additionally, computing power is scarce for Indonesian research institutions, which poses a challenge in accessibility and a barrier to adoption. We show that statistical approaches can be competitive in some cases while being significantly more efficient. Such approaches present an alternative in resource-constrained settings.
% Therefore, further study is needed to see their trade-off.

% Entries for the entire Anthology, followed by custom entries
\bibliography{custom}
\bibliographystyle{acl_natbib}

\clearpage
\appendix

\section{Data Statement for \datasetname{}}
\label{app:statement}
% Hapus link kalau sudah selesai
% GUIDELINES: \url{https://techpolicylab.uw.edu/data-statements/}

\subsection{General Information}

\paragraph{Dataset title} \datasetname{}

\paragraph{Dataset curators} Alham Fikri Aji (MBZUAI), Rahmad Mahendra (Universitas Indonesia), Samuel Cahyawijaya (HKUST), Ade Romadhony (Telkom University, Indonesia), Genta Indra Winata (Bloomberg), Fajri Koto (University of Melbourne), Kemal Kurniawan (University of Melbourne)

% \paragraph{Dataset curators} Redacted for anonymity

\paragraph{Dataset version} 1.0 (May 2022)

% \paragraph{Dataset citation} TODO

\paragraph{Data statement author} Kemal Kurniawan (University of Melbourne)

% \paragraph{Data statement author} Redacted for anonymity

\paragraph{Data statement version} 1.0 (February 2022)

% \paragraph{Data statement citation} Kurniawan, Kemal. (2022). \textit{Data Statement for \datasetname{}}. Version 1.0. University of Melbourne. TODO url.

\subsection{Executive Summary}

\datasetname{} is a multilingual parallel corpus across 10 local languages in Indonesia: Acehnese, Balinese, Banjarese, Buginese, Madurese, Minangkabau, Javanese, Ngaju, Sundanese, and Toba Batak. The data was translated obtained by human translation from Indonesian and human-assisted quality assurance. 

\subsection{Curation Rationale}

The goal of the dataset creation process is to provide gold-standard sentiment analysis
corpora for Indonesian local languages. The Indonesian data is sampled from SmSA \cite{purwarianti2019improving}, an Indonesian sentiment analysis corpus. SmSA is chosen among other corpora (e.g.,
HoASA \cite{azhar2019multi} based on (1) the agreement of our manual re-annotation of a small and randomly
selected samples and (2) manual inspection to ensure that the topics are diverse. After sampling,
the data is edited and/or filtered to remove harmful contents and maintain quality.
Several criteria are used in this process:
\begin{enumerate}[noitemsep]
    \item Is the sentiment label correct?
    \item Does the sentence contain multiple sentiments?
    \item Does the sentence contain harmful content that discriminates against race, religion, or other protected
    groups?
    \item Does the sentence contain an attack toward an individual or is abusive?
    \item Is the sentence politically charged?
    \item Is the sentence overly Bandung/Sunda-centric?\footnote{Bandung is the capital city of West Java, in which Sunda is the ethnic group.}
    \item Will the sentence be difficult to translate into local languages?
    \item Are there any misspellings?
\end{enumerate}

\subsection{Documentation for Source Datasets}

\datasetname{} is obtained by translating SmSA \cite{purwarianti2019improving}, an Indonesian sentiment analysis dataset. 

\subsection{Language Variety}

\datasetname{} covers a total of 10 local languages spoken in Indonesia (ID) as shown in Table~\ref{tab:lang-variety}. 
%TODO describe the scripts used. TODO add 'as spoken in' column?

\begin{table*}[!ht]
    \centering
       \resizebox{0.95\linewidth}{!}{
    \begin{tabular}{@{}lcll@{}}
    \toprule
        \textbf{Language} & \textbf{ISO 639-3} & \textbf{Annotators' Dialect} & \textbf{Example} \\
    \midrule
        Acehnese & ace & Banda Aceh & Meureutoh rumoh di Medan keunong ie raya\\
        Balinese & ban & Lowland & Satusan umah ring medan merendem banjir\\
        Toba Batak & bbc & Toba, Humbang & Marratus jabu di medan na hona banji\\
        Banjarese & bjn & Hulu, Kuala & Ratusan rumah di medan tarandam banjir \\
        Buginese & bug & Sidrap & Maddatu bola okko medan nala lempe \\
        Javanese & jav & Matraman & Atusan omah ing medan kebanjiran\\
        Madurese & mad & Situbondo & Ratosan bangko e medan tarendem banjir \\
        Minangkabau & min & Padang, Agam & Ratuihan rumah di medan tarandam banjir \\
        Ngaju & nij & Kapuas, Kahayan & Ratusan huma hong medan lelep awi banjir\\
        Sundanese & sun & Priangan & Ratusan bumi di medan karendem banjir\\
    \bottomrule
    \end{tabular}
      }
    \caption{Local languages spoken in Indonesia (ID) that are covered in \datasetname{}.} %Number of speakers column represents the speakers of the whole language (not just that particular dialect).}
    \label{tab:lang-variety}
\end{table*}

%\begin{table}[!t]
%    \centering
 
%    \begin{tabular}{@{}ll@{}}
%    \toprule
%        Language & Text\\
%    \midrule
%        Indonesian & Ratusan rumah di medan terendam banjir \\ %\midrule
        %\texttt{Translations} \\
        %\midrule
%        English & Hundreds of houses in Medan were submerged \\ 
%        & by the flood \\
%        Acehnese &  \\
%        Balinese &  \\
%        Banjarese &  \\
%        Buginese &  \\
%        Javanese &  \\ 
%        Madurese &  \\
%        Minangkabau &  \\
%        Ngaju & \\ 
%        Sundanese &  \\
%        Toba Batak &  \\ \bottomrule
%    \end{tabular}
  
%    \caption{Example of the NusaX data.}
%    \label{tab:nusaxexample}
%\end{table}

\subsection{Speaker Demographic}
The SmSA dataset was obtained from social media and online forums: Twitter, Zomato,
TripAdvisor, Facebook, Instagram, Qraved. We can assume the users' age ranges from 25 to 34 years, which is the age range of the majority of Indonesian social media users\footnote{\scriptsize \url{https://www.statista.com/statistics/997297/indonesia-breakdown-social-media-users-age-gender/}}.
%TODO describe what's reported in SmSA paper, and perhaps add our own judgment if missing

\subsection{Annotator Demographic}

A total of 28 translators are employed in the translation process. All translators are
Indonesian and recruited by via either online surveys or personal contacts. They are
then selected based on (1) the self-reported fluency in the local language into which they
would be translating and (2) the highest education level achieved. Those who (a) are native
speakers of or fluent in the target local language and (b) finished at least high school
education (id: \textit{SMA/sederajat}) are selected.

\paragraph{Acehnese} There are 3 translators for Acehnese, but only 2 of them responded
when asked for demographic information. Thus, what follows is the demographic information
of only those 2 translators. One has some experience in translation work, while the other
does not. One identifies as male, and the other as female. Both are in their 20s. Lastly,
one works as a freelancer, while the other is a farmer.

\paragraph{Balinese} Three people translate into Balinese. Two of them have previous experience
in translation work, and both identify as female. The other one, who identifies as male,
does not have such experience. Two of them are aged 20-29 years old, while the other is in
their 30s. Their occupations are university lecturer, school teacher, and civil employee respectively.

\paragraph{Banjarese} Two translators are employed for Banjarese, but only one responded
when asked for demographic information. The translator has prior experience in translation
work, identifies as male, is in his 40s, and works as a university lecturer.

\paragraph{Buginese} Buginese is translated by 2 people, but only one responded when asked
for demographic information. The person has prior translation experience,
identifies as male, is aged 30-39 years old, and runs an Islamic boarding school as a living.

\paragraph{Javanese} Four translators are employed for Javanese, but one did not respond
when asked for demographic information. The other three have prior experience in
translation work. Among them, two identify as female, and one as male. All of them are
in their 20s. Two of them are university students, and the other one works as a freelance
assistant editor.

\paragraph{Madurese} There are 3 translators for Madurese. Only one of them has previous
experience in translation work. Two of them identify as female, while the other as male.
One person is aged under 20 years old and is a university student. The others are 20-29
years old and work as a school teacher and an employee in a private company respectively.

\paragraph{Minangkabau} Three people translate into Minangkabau. Two of them have previous
translation experience. All three identify as female and are aged 20-29 years old. They work
as a civil employee, a university student, and a senior data annotator respectively.

\paragraph{Ngaju} Two translators work on Ngaju, but only one responded when asked for
demographic information. The translator has prior experience, identifies as female, is
aged no less than 50 years old, and is a stay-at-home mother.

\paragraph{Sundanese} There are 5 translators for Sundanese, four of which identify as female, and
the other one as male. Three translators are in their 20s, one is younger than 20 years old,
and the remaining one is in their 30s. The translators work as a school teacher, a university
student, a university lecturer, and the remaining two as employees in a private company.

\paragraph{Toba Batak} Three translators are employed for Toba Batak. One has prior
translation experience. Two translators identify as male while the other as female.
All three are in their 20s. One works for a private company, and the others are university students.

\section{Hyperparameters} \label{app:hyperparameters}

\subsection{Sentiment Analysis}

\begin{table}[!ht]
    \centering
     \resizebox{\linewidth}{!}{
     \small
    \begin{tabular}{lccc}
    \toprule
        \textbf{Hyperparams} & \textbf{NB} & \textbf{SVM} & \textbf{LR} \\
    \midrule
        feature & \{BoW, tfidf\}  & \{BoW, tfidf\}  & \{BoW, tfidf\} \\
        alpha & (0.001 - 1) & -- & --\\
        C & -- & (0.01 - 100) & (0.001 - 100) \\
        kernel & -- & \{rbf, linear\} & -- \\
    \bottomrule
    \end{tabular}
    }
    \caption{Hyperparameters of statistical models on sentiment analysis.
    \label{tab:classic-range}
    }
\end{table}

\begin{table}[!ht]
    \centering
    \resizebox{\linewidth}{!}{
        % \small
        \begin{tabular}{lc}
        \toprule
            \textbf{Hyperparams} & \textbf{Values} \\
        \midrule
            learning rate & [1e-4, 5e-5, \textbf{1e-5}, 5e-6, 1e-6]  \\
            batch size & [4, 8, 16, \textbf{32}] \\
            num epochs & 100 \\
            early stop & 3 \\
            max norm & 10 \\
            optimizer & Adam \\
            \hspace{3mm} Adam $\beta$ & (0.9, 0.999) \\
            \hspace{3mm} Adam $\gamma$ & 0.9 \\
            \hspace{3mm} Adam $\epsilon$ & 1e-8 \\
        \bottomrule
        \end{tabular}
    }
    \caption{Hyperparameters of pre-trained LMs on sentiment analysis. \textbf{Bold} denotes the best hyperparameter setting.}
    \label{tab:deep-classification-range}
\end{table}

For statistical models, we use a spaCy as our toolkit, and we perform grid-search over the parameter ranges shown in Table~\ref{tab:classic-range} and select the best performing model over the devset. For all pre-trained LMs, we perform grid-search over batch size and learning rate while keeping the other hyperparameters fixed. The list of hyperparameters is shown in Table~\ref{tab:deep-classification-range}.

\subsection{Machine Translation}
Table~\ref{tab:hyperparams-mt-neural} shows the hyperparameters of deep learning models on machine translation. 

\begin{table}[!ht]
    \centering
    \resizebox{\linewidth}{!}{
    \begin{tabular}{@{}lcccc@{}}
    \toprule
        \textbf{Hyperparams} & \textbf{IndoGPT} & \textbf{IndoBARTv2} & \textbf{mBART-50} & \textbf{mT5}$_{\textbf{BASE}}$\\
    \midrule
        learning rate & 1e-4 & 1e-4 & 2e-5 & 5e-4 \\
        batch size & \multicolumn{4}{c}{$16$}  \\
        gamma & 0.98 & 0.98 & 0.98 & 0.95 \\
        max epochs & \multicolumn{4}{c}{$20$}  \\
        early stop & \multicolumn{4}{c}{$10$} \\
        seed & \multicolumn{4}{c}{$\{1...5\}$} \\
    \bottomrule
    \end{tabular}
    }
    \caption{Hyperparameters of pretrained LMs on machine translation.}
    \label{tab:hyperparams-mt-neural}
\end{table}

\section{Dataset Statistics}
\label{app:statistics}

% \begin{table}[!ht]
%     \centering
%      \resizebox{0.8\linewidth}{!}{
%      \small
%     \begin{tabular}{lccc}
%     \toprule
%         \textbf{Subset} & \textbf{Negative} & \textbf{Neutral} & \textbf{Positive} \\
%     \midrule
%         Train & 192 & 119 & 189 \\
%         Valid & 38 & 24 & 38 \\
%         Test & 153 & 96 & 151 \\
%     \bottomrule
%     \end{tabular}
%     }
%     \caption{Label distribution of NusaX Sentiment dataset.
%     \label{tab:label-distribution}
%     }
% \end{table}

% In this section, we present more detail statistics of our NusaX datasets. For the NusaX sentiment dataset, each language has the same label distribution and we show the label distribution of each dataset subset on Table~\ref{tab:label-distribution}. The label ratio is maintained on each dataset subset to ensure a similar distribution between each subset. 

\begin{figure*}[!ht]
    \centering
    \begin{minipage}{.32\linewidth}
        \centering
        \begingroup
        \includegraphics[width=\linewidth]{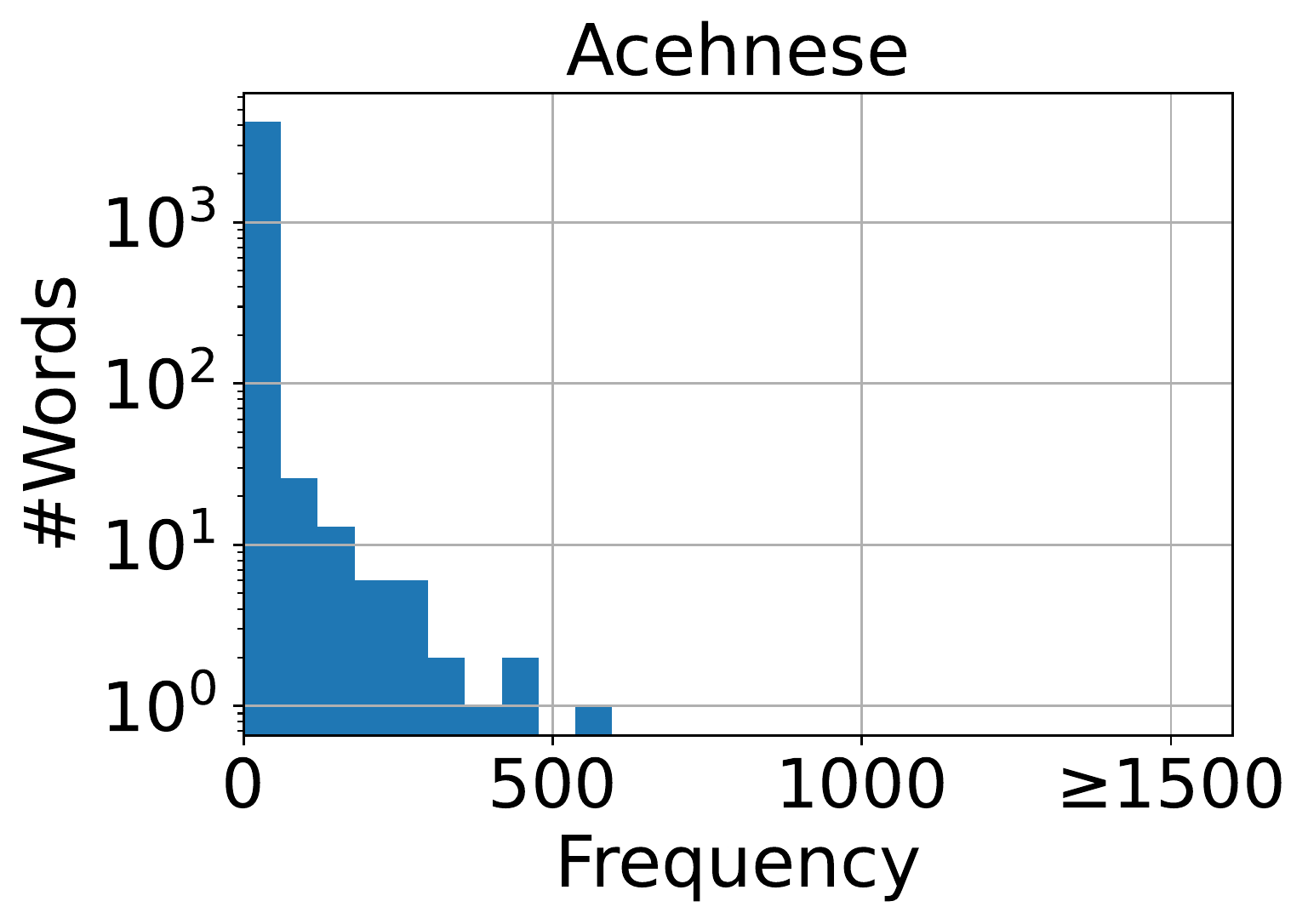}
        \endgroup
    \end{minipage}%
    \hspace{5pt}
    \begin{minipage}{.32\linewidth}
        \centering
        \begingroup
        \includegraphics[width=\linewidth]{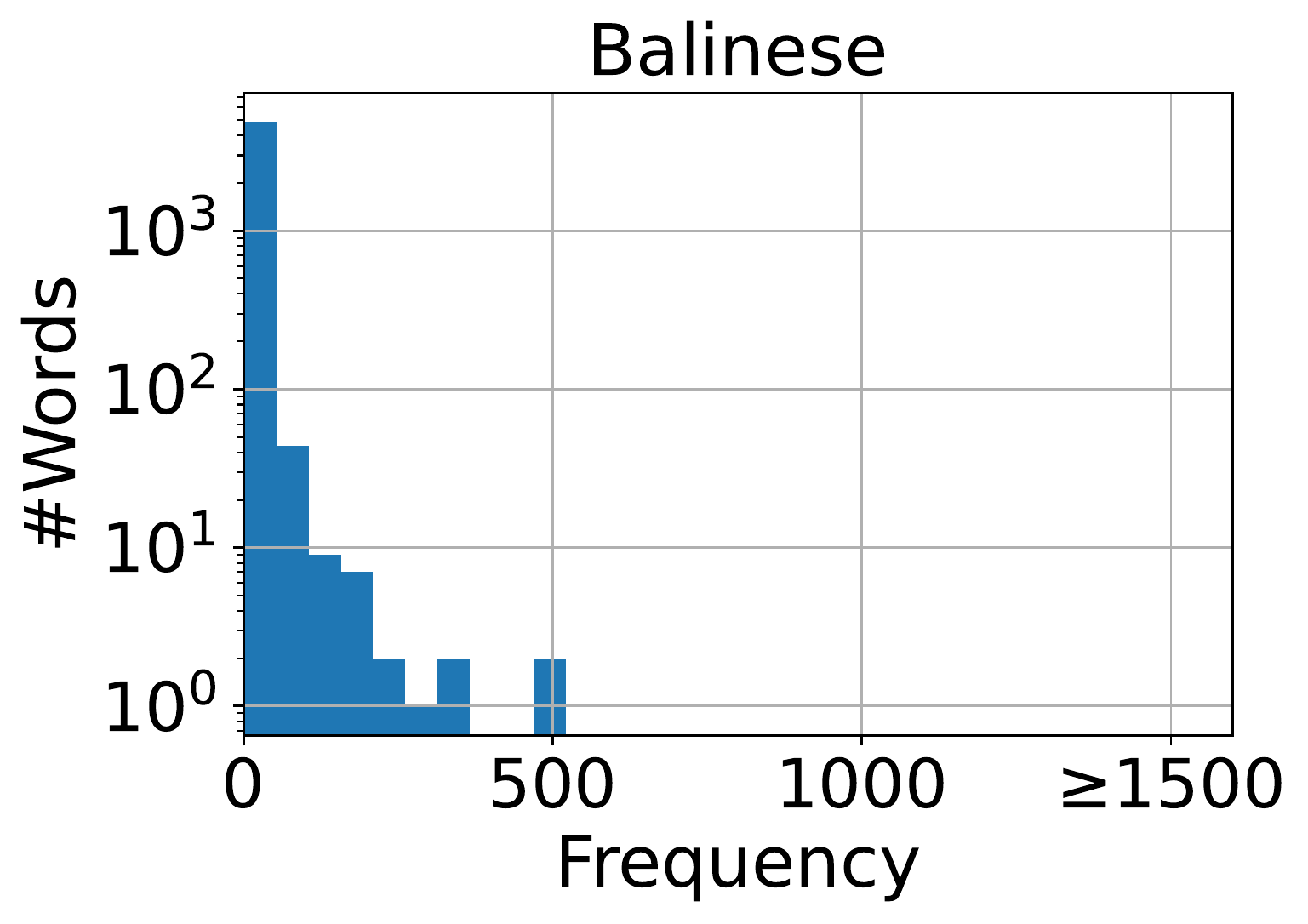}
        \endgroup
    \end{minipage}
    \begin{minipage}{.32\linewidth}
        \centering
        \begingroup
        \includegraphics[width=\linewidth]{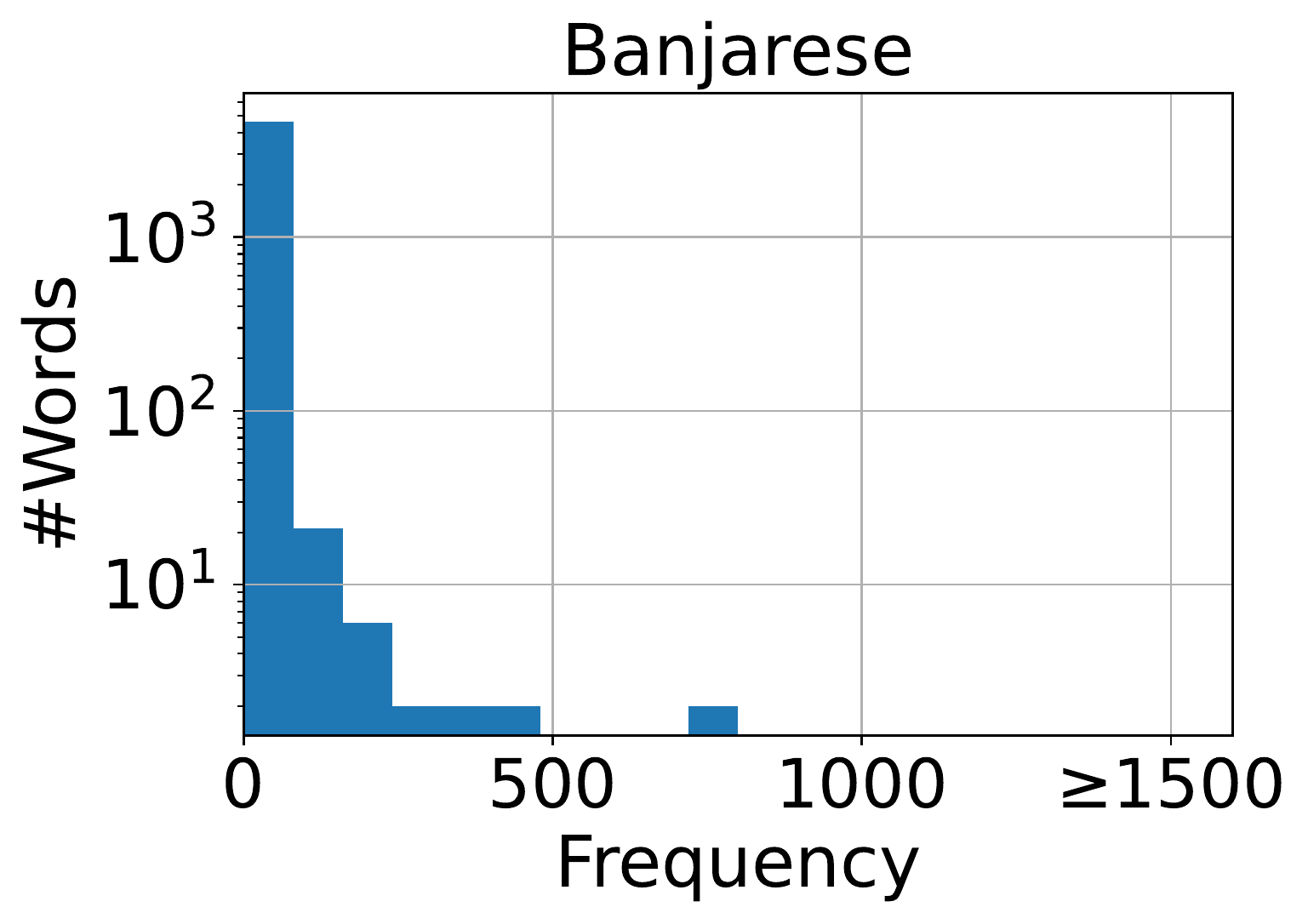}
        \endgroup
    \end{minipage}%
    \hspace{5pt}
    \begin{minipage}{.32\linewidth}
        \centering
        \begingroup
        \includegraphics[width=\linewidth]{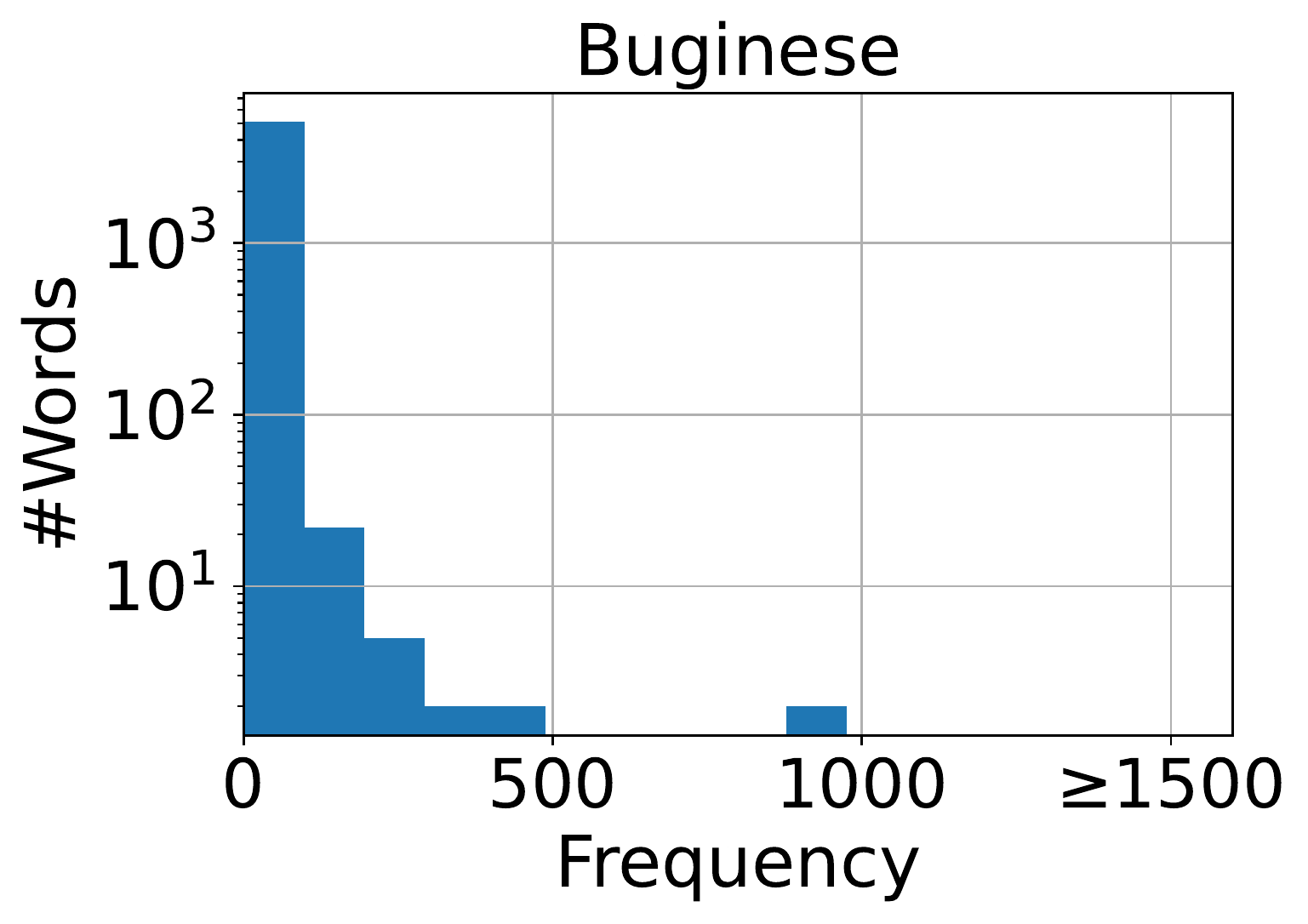}
        \endgroup
    \end{minipage}
    \begin{minipage}{.32\linewidth}
        \centering
        \begingroup
        \includegraphics[width=\linewidth]{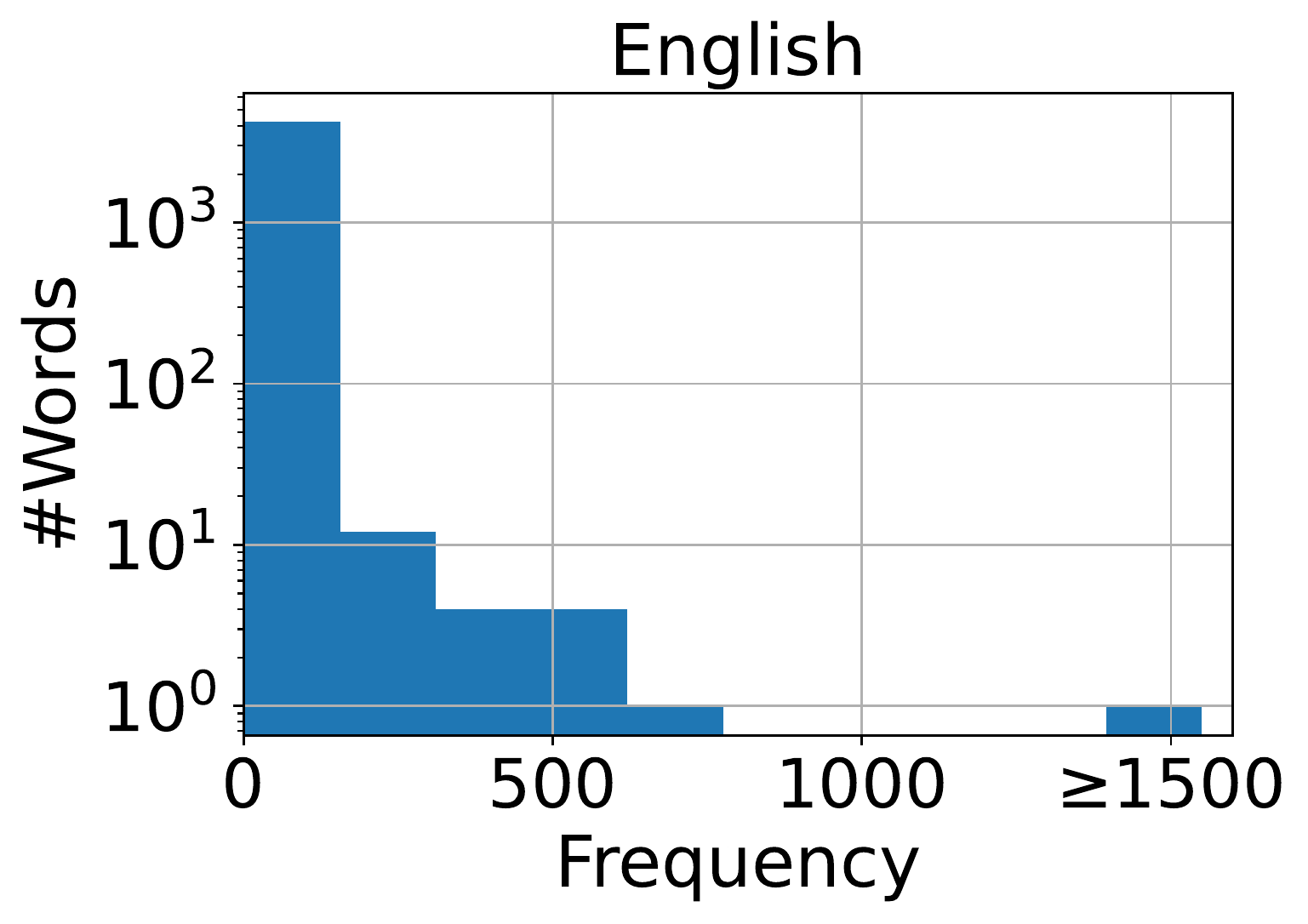}
        \endgroup
    \end{minipage}%
    \hspace{5pt}
    \begin{minipage}{.32\linewidth}
        \centering
        \begingroup
        \includegraphics[width=\linewidth]{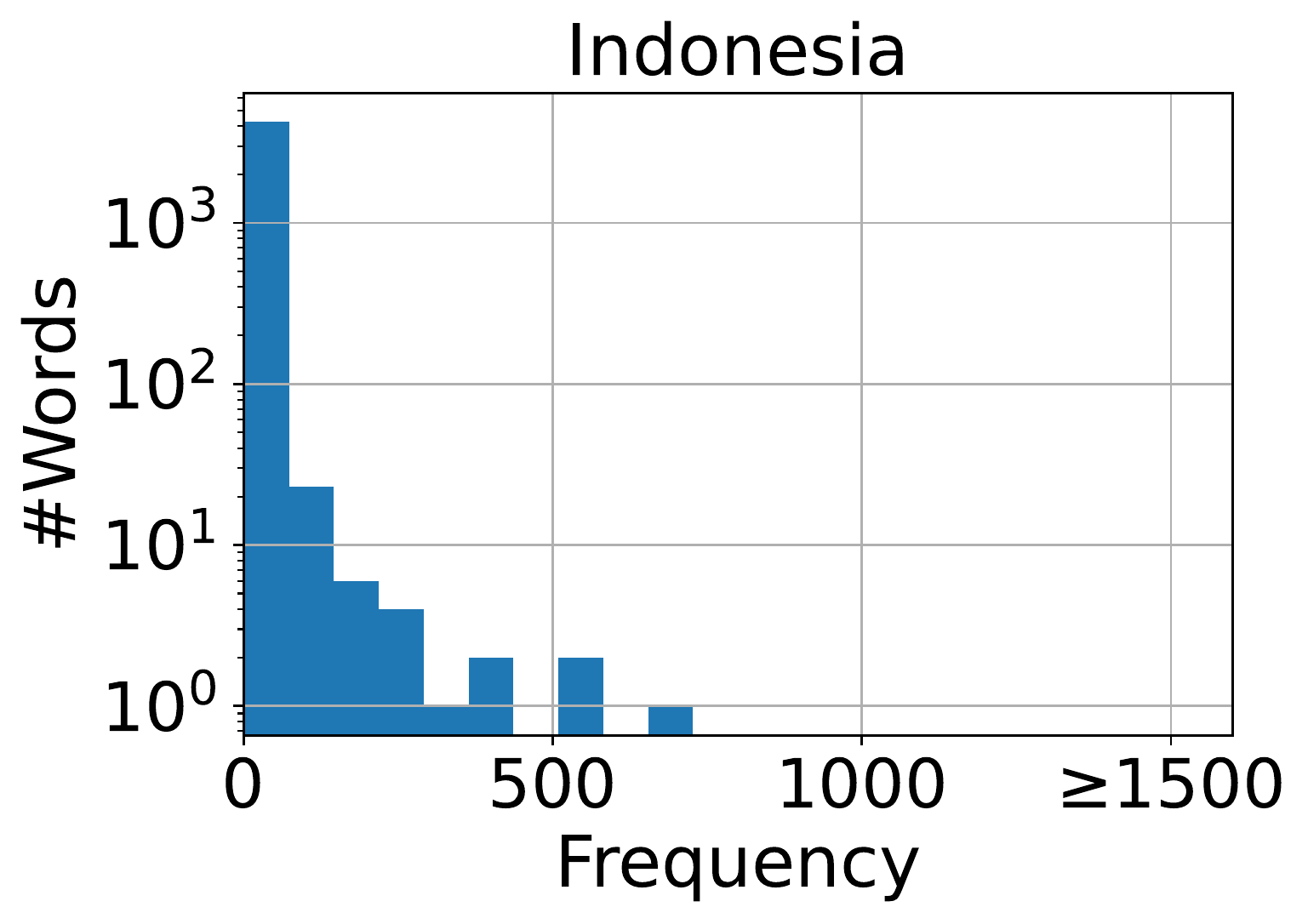}
        \endgroup
    \end{minipage}
    \begin{minipage}{.32\linewidth}
        \centering
        \begingroup
        \includegraphics[width=\linewidth]{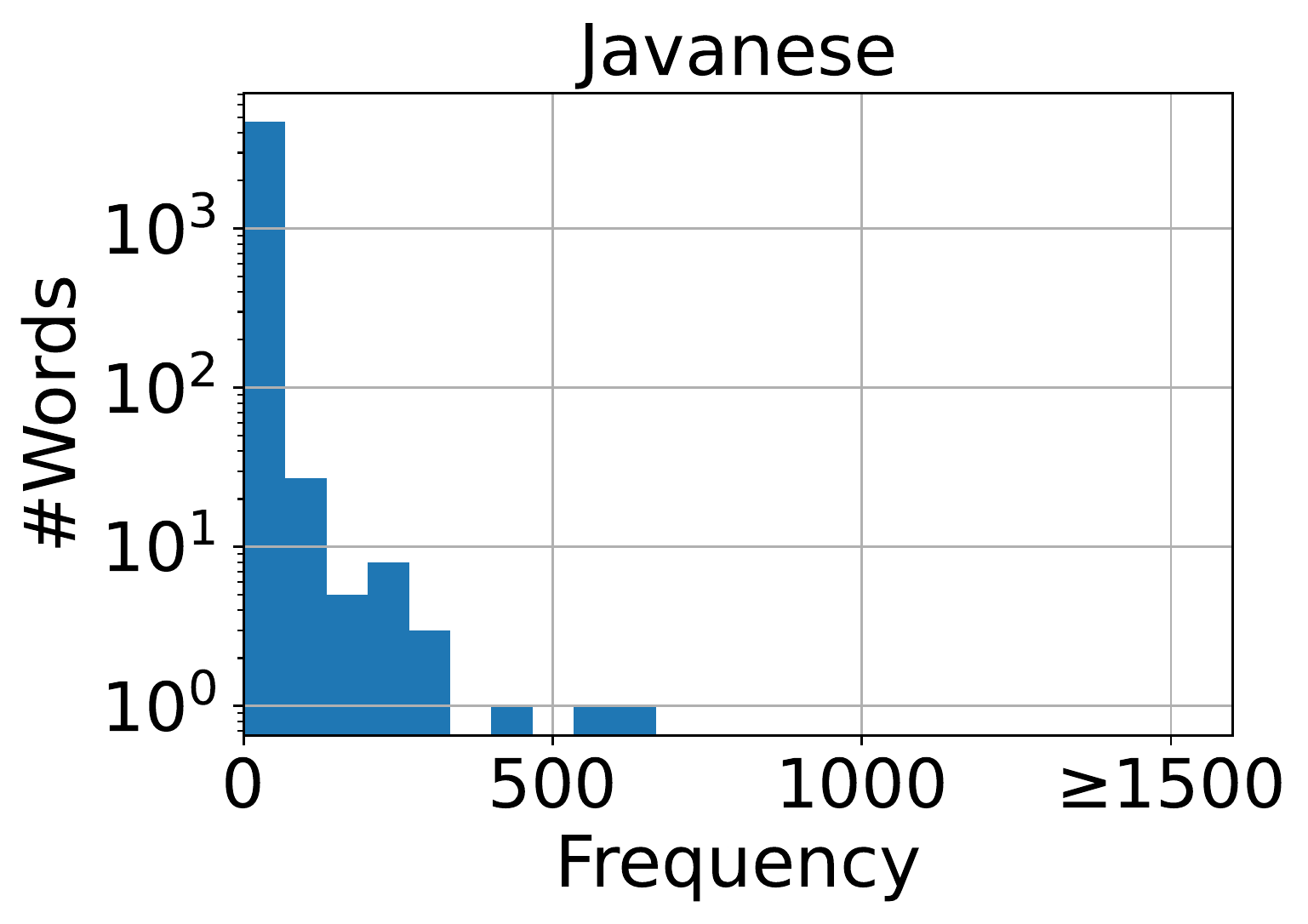}
        \endgroup
    \end{minipage}%
    \hspace{5pt}
    \begin{minipage}{.32\linewidth}
        \centering
        \begingroup
        \includegraphics[width=\linewidth]{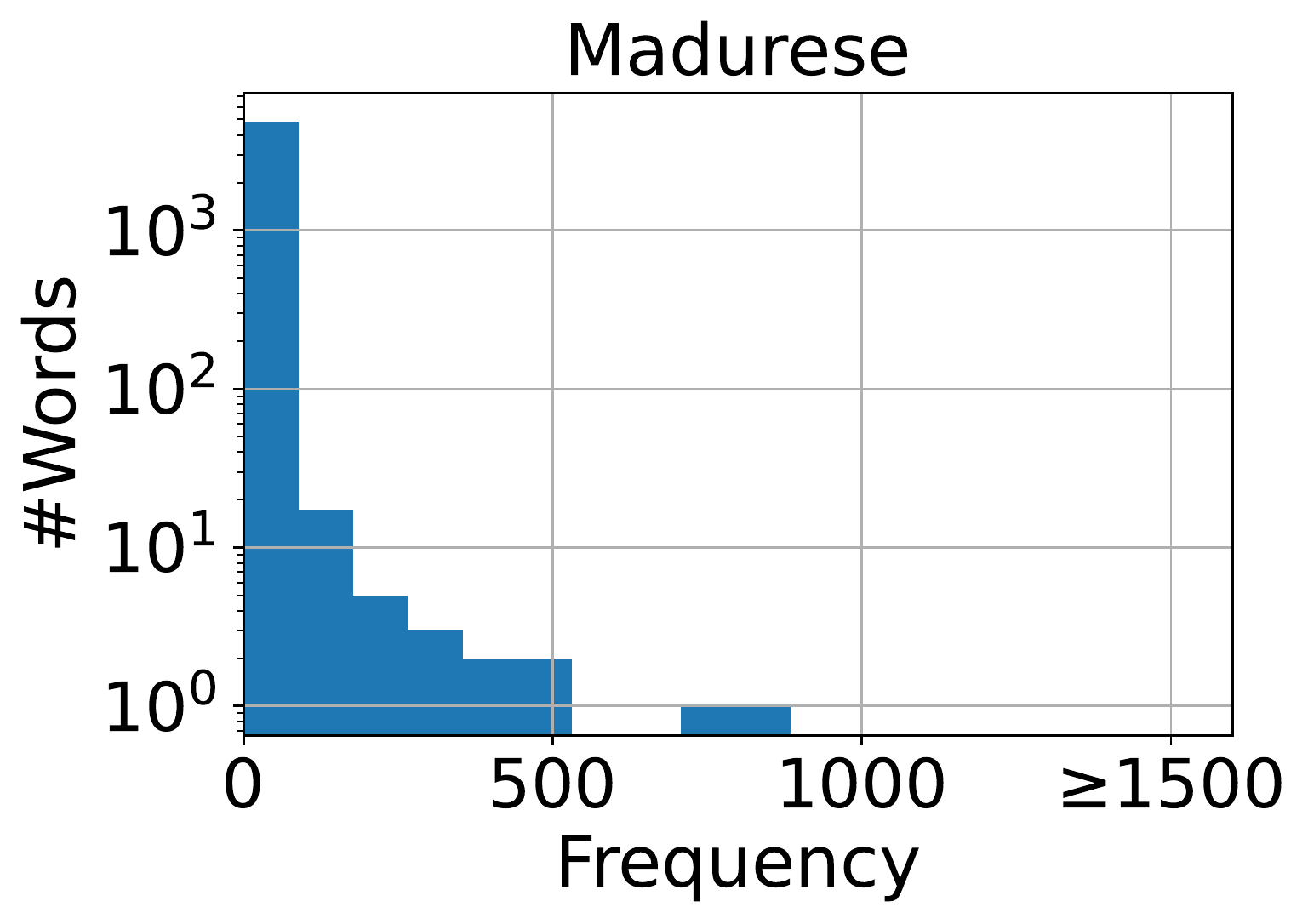}
        \endgroup
    \end{minipage}
    \begin{minipage}{.32\linewidth}
        \centering
        \begingroup
        \includegraphics[width=\linewidth]{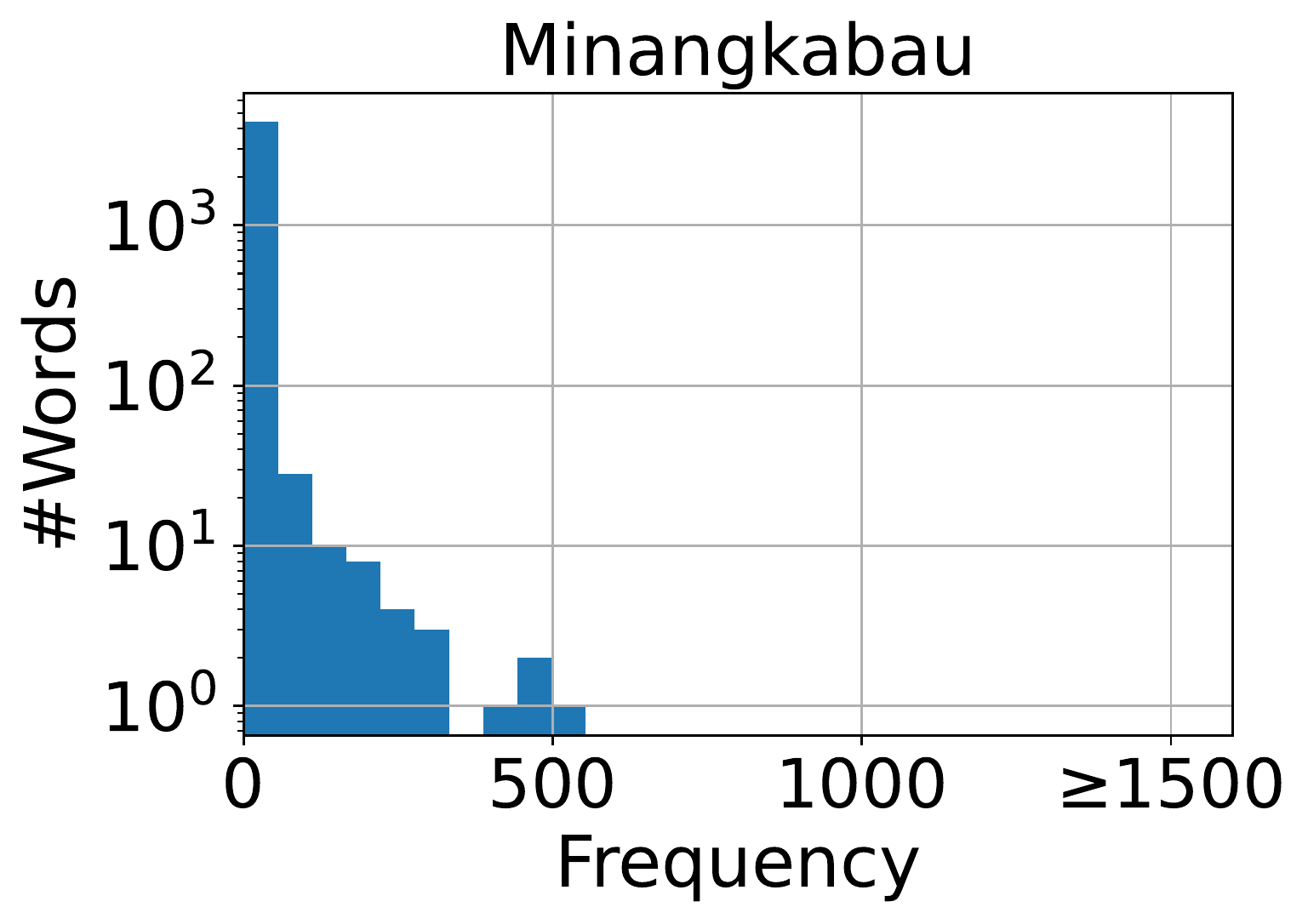}
        \endgroup
    \end{minipage}%
    \hspace{5pt}
    \begin{minipage}{.32\linewidth}
        \centering
        \begingroup
        \includegraphics[width=\linewidth]{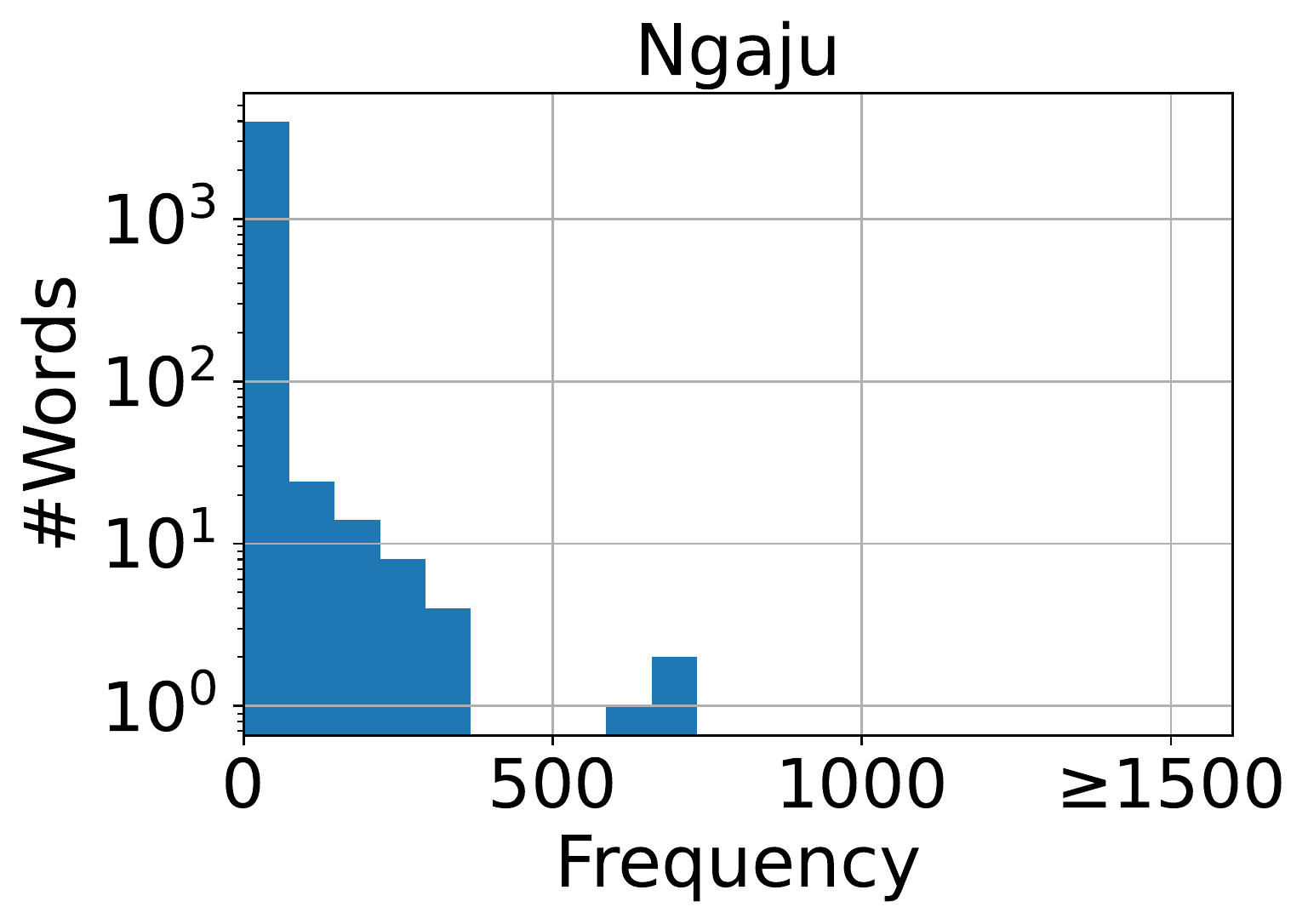}
        \endgroup
    \end{minipage}
    \begin{minipage}{.32\linewidth}
        \centering
        \begingroup
        \includegraphics[width=\linewidth]{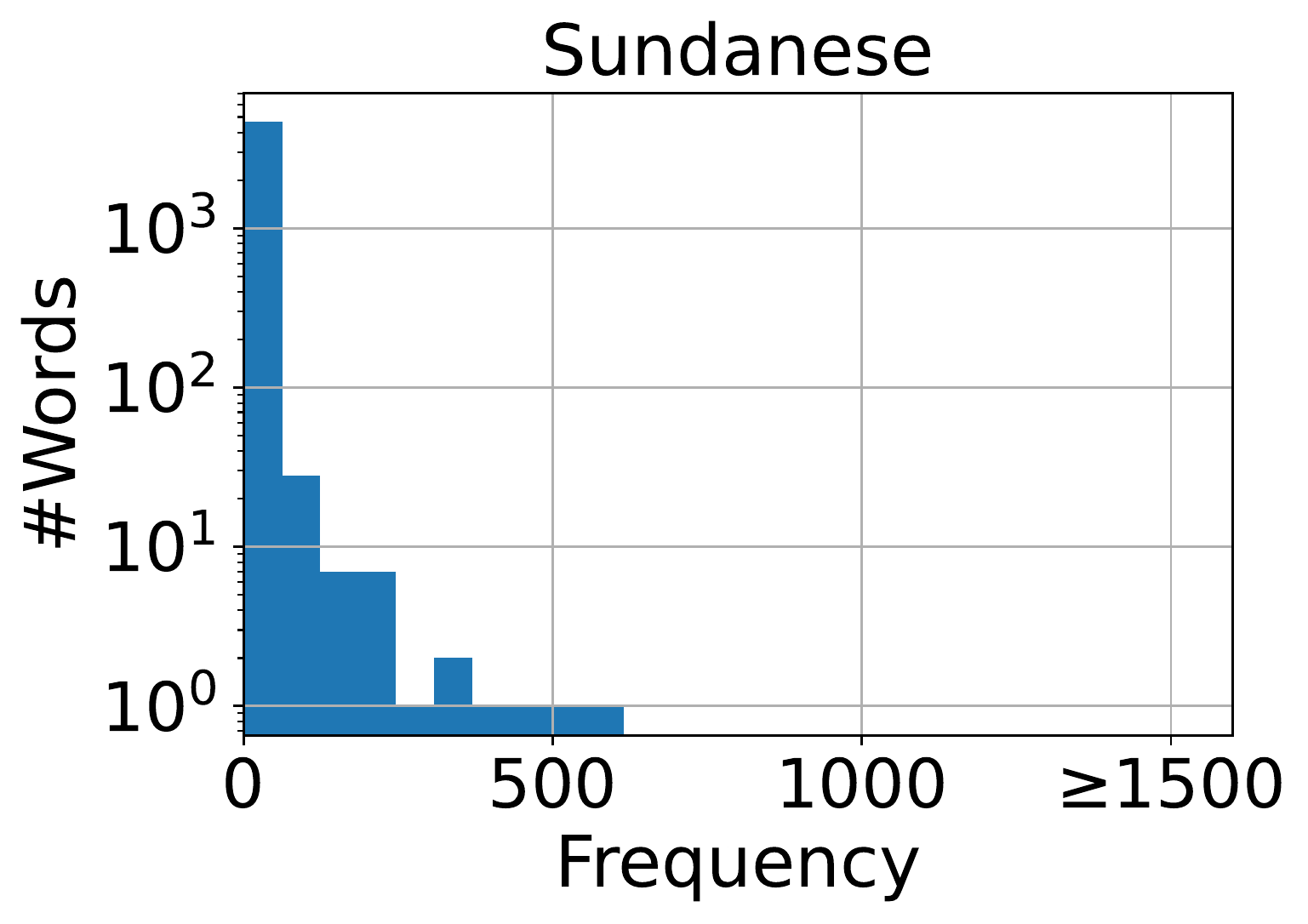}
        \endgroup
    \end{minipage}%
    \hspace{5pt}
    \begin{minipage}{.32\linewidth}
        \centering
        \begingroup
        \includegraphics[width=\linewidth]{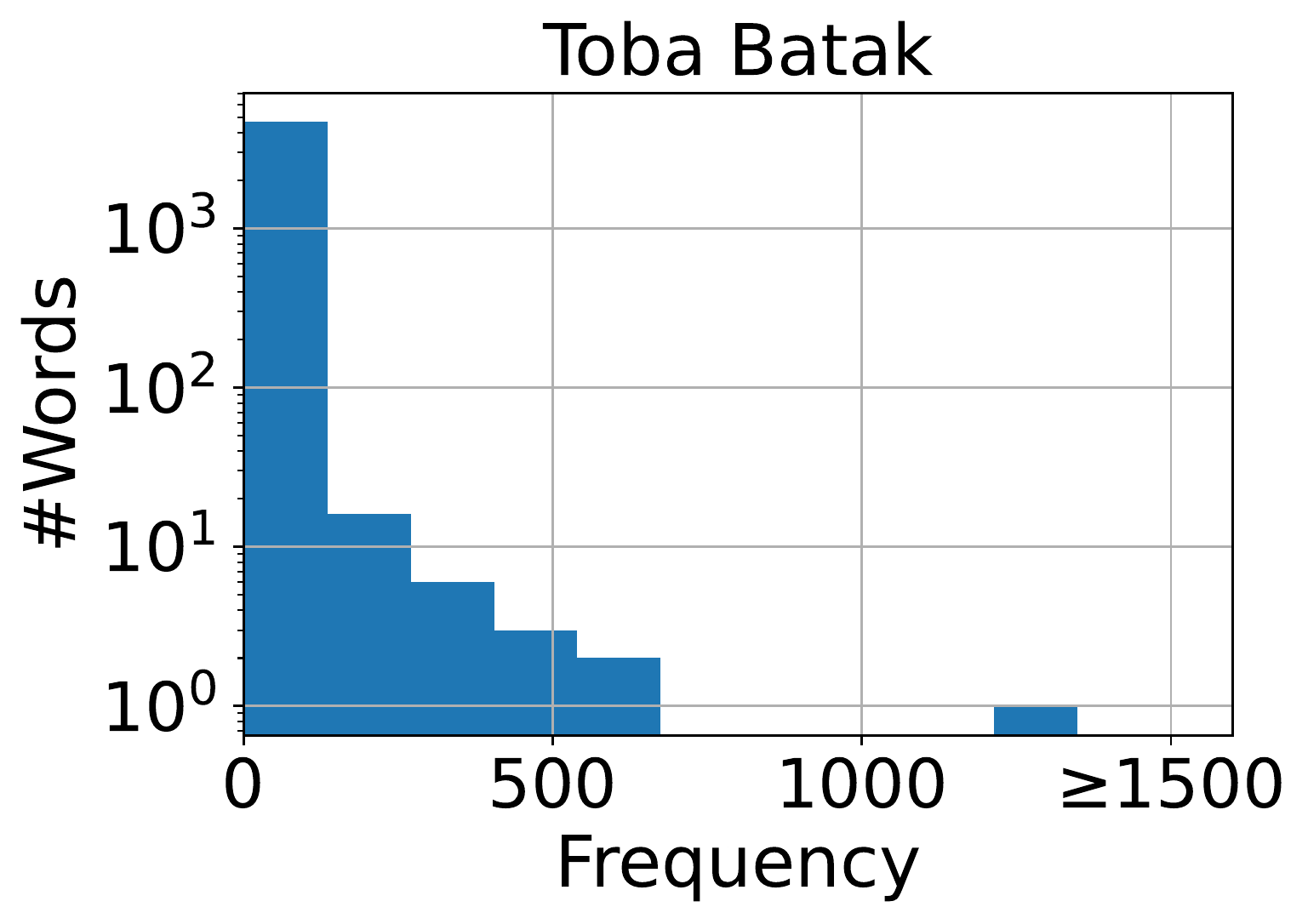}
        \endgroup
    \end{minipage}
    \caption{Word frequency histogram for each language in NusaX.}
    \label{fig:word-histogram}
\end{figure*}

In this section, we present more detail statistics of our NusaX datasets. To evaluate the difference between each language in the NusaX dataset, we analyze the vocabulary characteristic for each language. We collect the vocabulary for each language by removing all the punctuation in the sentence and tokenize the sentence with the spaCy tokenizer.~\footnote{\url{https://github.com/explosion/spaCy}} We show the vocabulary size and the top-10 words for each language on Table~\ref{tab:top-10-words}, and the vocabulary histogram for each language in Figure~\ref{fig:word-histogram}. We can see that the most common words between Indonesian and other local languages vary a lot, despite having a similar vocabulary size and histogram pattern. This shows the intuitive difference between Indonesian and local languages in Indonesia.

\begin{figure*}[!t]
    \centering
    \includegraphics[width=\linewidth]{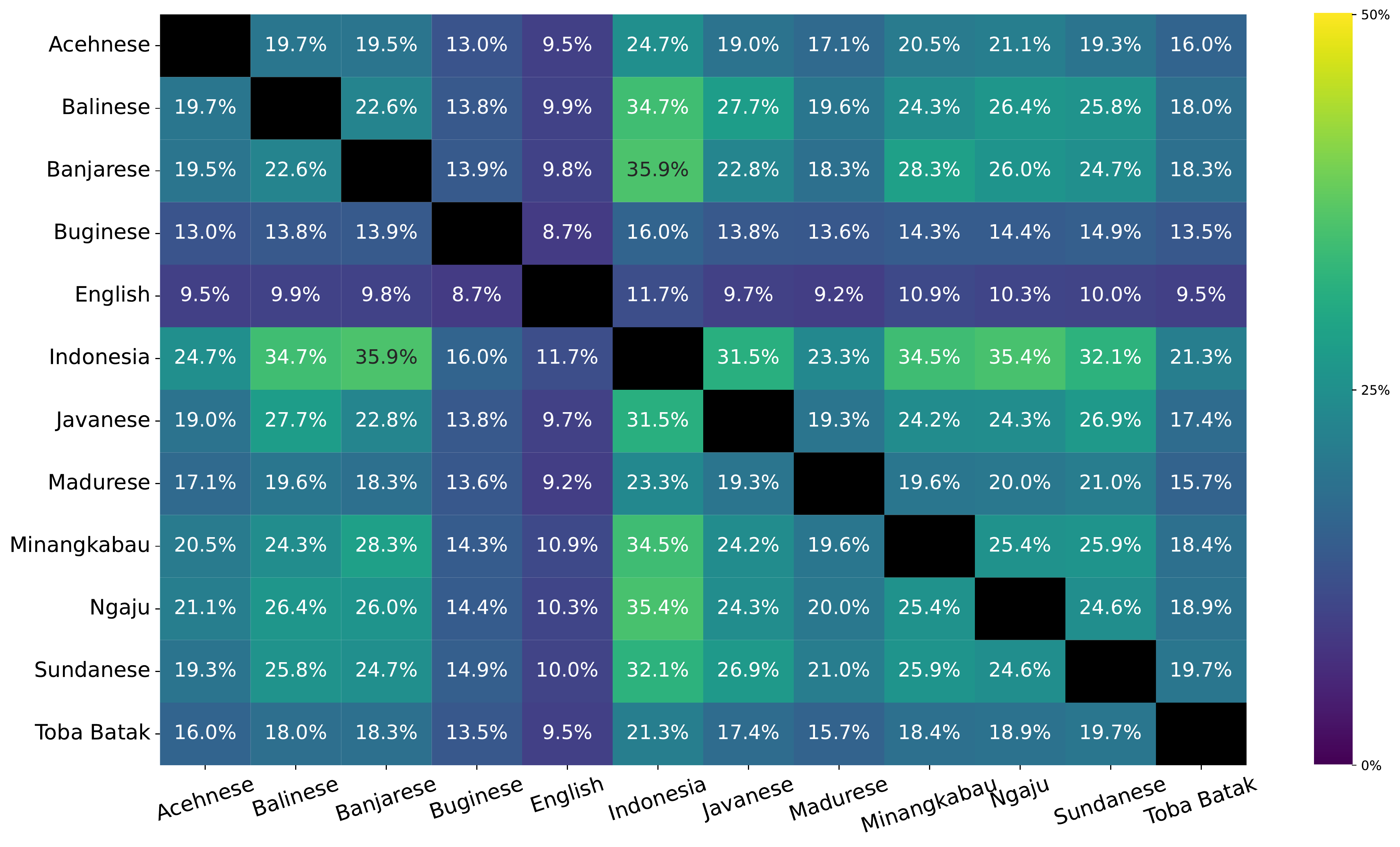}
    \caption{Vocabulary overlap between language pairs in NusaX dataset.}
    \label{fig:vocab-overlap}
\end{figure*}

We further measure the vocabulary overlap over different language pairs. We measure the vocabulary overlap for each pair of languages by measuring the intersection over union (IoU) of the two vocabularies. We show the vocabulary overlap in Figure~\ref{fig:vocab-overlap}. From the results, we can conclude that English has the smallest vocabulary overlap with the other languages. This makes sense since English comes from a different language family, i.e., Indo-European language under the Germanic language branch, while the others are from the Austronesian language family under the Malayo-Polynesian branch. Other languages that have low vocabulary overlap are Buginese (bug) and Toba Batak (bbc). This aligns with our discussion in \S\ref{sec:results}, which shows the distinction between these languages and the other languages in the NusaX dataset.

\begin{table}[!t]
    \centering
     \resizebox{0.955\linewidth}{!}{
     \small
    \begin{tabular}{ccc}
        \toprule
        \textbf{Toba Batak} & \textbf{Balinese} & \textbf{Banjarese} \\ 
        % & \textbf{Minangkabau} & \textbf{English} & \textbf{Ngaju} \\
        \footnotesize{\textbf{(4681 words)}} & \footnotesize{\textbf{(4927 words)}} & \footnotesize{\textbf{4631 words}} \\ 
        % & \footnotesize{\textbf{(446 words)}} & \footnotesize{\textbf{(4233 words)}} & \footnotesize{\textbf{(4005 words)}} \\
        \midrule
        na & lan & nang \\
        % & di & the & te \\
        di & sane & wan \\ 
        % & nan & and & dengan \\
        dohot & ring & di \\
        % & dan & to & ji \\
        ni & ane & kada \\
        % & jo & is & mangat \\
        tu & ne & ulun \\
        % & untuak & a & eka \\
        do & sajan & nyaman \\
        % & awak & of & akan \\
        dang & tiyang & gasan \\
        % & yang & for & aku \\
        pe & tiang & banar \\
        % & lamak & in & jadi \\
        tabo & ajak & makan \\ 
        \midrule \\
        % & ka & i & diak \\
        \midrule
        
        % \textbf{Toba Batak} & \textbf{Balinese} & \textbf{Banjarese} & 
        \textbf{Minangkabau} & \textbf{English} & \textbf{Ngaju} \\
        % \footnotesize{\textbf{(4681 words)}} & \footnotesize{\textbf{(4927 words)}} & \footnotesize{\textbf{4631 words}} &
        \footnotesize{\textbf{(446 words)}} & \footnotesize{\textbf{(4233 words)}} & \footnotesize{\textbf{(4005 words)}} \\
        \midrule
        % na & lan & nang & 
        di & the & te \\
        % di & sane & wan & 
        nan & and & dengan \\
        % dohot & ring & di & 
        dan & to & ji \\
        % ni & ane & kada & 
        jo & is & mangat \\
        % tu & ne & ulun & 
        untuak & a & eka \\
        % do & sajan & nyaman & 
        awak & of & akan \\
        % dang & tiyang & gasan & 
        yang & for & aku \\
        % pe & tiang & banar & 
        lamak & in & jadi \\
        % tabo & ajak & makan & 
        ka & I & diak \\ 
        \midrule \\
        \midrule
        \textbf{Sundanese} & \textbf{Buginese} & \textbf{Indonesian} \\
        % & \textbf{Javanese} & \textbf{Acehnese} & \textbf{Maduranese} \\
        \footnotesize{\textbf{(4693 words)}} & \footnotesize{\textbf{(5118 words)}} & \footnotesize{\textbf{(4269 words)}} \\
        % & \footnotesize{\textbf{(4719 words)}} & \footnotesize{\textbf{(4250 words)}} & \footnotesize{\textbf{(4846 words)}} \\
        \midrule
        nu & e & yang \\ 
        % & sing & nyang & se \\
        sareng & na & di \\
        % & lan & ngon & e \\
        di & okko & dan \\
        % & ora & hana & bik \\
        teu & sibawa & tidak \\
        % & karo & lon & engkok \\
        pisan & iya & saya \\
        % & aku & that & ben \\
        abdi & de & dengan \\
        % & ing & mangat & tak \\
        ka & i & ini \\
        % & iki & nyoe & nyaman \\
        ieu & ko & enak \\
        % & ning & dan & ka \\
        aya & ladde & untuk \\ 
        \midrule \\
        % & enak & bak & ghebey \\
        \midrule
        % \textbf{Sundanese} & \textbf{Buginese} & \textbf{Indonesian} &
        \textbf{Javanese} & \textbf{Acehnese} & \textbf{Maduranese} \\
        % \footnotesize{\textbf{(4693 words)}} & \footnotesize{\textbf{(5118 words)}} & \footnotesize{\textbf{(4269 words)}} &
        \footnotesize{\textbf{(4719 words)}} & \footnotesize{\textbf{(4250 words)}} & \footnotesize{\textbf{(4846 words)}} \\
        \midrule
        % nu & e & yang & 
        sing & nyang & se \\
        % sareng & na & di & 
        lan & ngon & e \\
        % di & okko & dan & 
        ora & hana & bik \\
        % teu & sibawa & tidak & 
        karo & lon & engkok \\
        % pisan & iya & saya & 
        aku & that & ben \\
        % abdi & de & dengan & 
        ing & mangat & tak \\
        % ka & i & ini & 
        iki & nyoe & nyaman \\
        % ieu & ko & enak & 
        ning & dan & ka \\
        % aya & ladde & untuk & 
        enak & bak & ghebey \\
        \bottomrule
    \end{tabular}
    }
    \caption{Vocabulary size (in bracket) and top-10 words on each language in the NusaX dataset.}
    \label{tab:top-10-words}
\end{table}

% \section{Examples}
% \label{app:examples}

% In Table~\ref{tab:example}, we provide several examples of the translated parallel data of NusaX corpus.

% \begin{table}[!ht]
%     \centering
%     \resizebox{\linewidth}{!}{
%     \begin{tabular}{@{}ll@{}}
%     \toprule
%         Language & Text\\
%     \midrule
%         Indonesian & Ratusan rumah di medan terendam banjir \\ \midrule
%         \texttt{Translations} \\
%         \midrule
%         English & Hundreds of houses in Medan were submerged \\ 
%         & by the flood \\
%         Acehnese & Meureutoh rumoh di Medan keunong ie raya \\
%         Balinese & Satusan umah ring medan merendem banjir \\
%         Banjarese & Ratusan rumah di medan tarandam banjir \\
%         Buginese & Maddatu bola okko medan nala lempe \\
%         Javanese & Atusan omah ing medan kebanjiran \\ 
%         Madurese & Ratosan bangko e medan tarendem banjir \\
%         Minangkabau & Ratuihan rumah di medan tarandam banjir \\
%         Ngaju & Ratusan huma hong medan lelep awi banjir\\ 
%         Sundanese & Ratusan bumi di medan karendem banjir \\
%         Toba Batak & Marratus jabu di medan na hona banji \\ \bottomrule
%     \end{tabular}
%     }
%     \caption{Example of the translation data.}
%     \label{tab:example}
% \end{table}

\end{document}